%% file: 0_main.tex
\documentclass[10pt,twocolumn,letterpaper]{article}

\usepackage[pagenumbers]{cvpr} %

\input{0_preamble}

\def\sepappendix{0}

\title{Lost in Translation, Found in Context: \\ Sign Language Translation with Contextual Cues}

\input{authors}

\begin{document}

\maketitle

\input{0_abstract}
\input{1_intro}

\input{2_related}
\input{3_method}
\input{4_experiments}
\input{5_conclusion}

\input{6_acknowledgements}

{
    \small
    \bibliographystyle{ieeenat_fullname}
    \bibliography{shortstrings, references, vgg_local}
}

\clearpage
{\noindent \large \bf {APPENDIX}}\\
\input{7_appendix}

\end{document}

%% file: 0_preamble.tex
\usepackage[dvipsnames]{xcolor}
\usepackage{booktabs} %
\usepackage{pifont} %

\usepackage{colortbl}
\usepackage{multirow}
\usepackage{multicol}
\usepackage{listings} %
\usepackage{titletoc} %
\usepackage{fancyvrb}
\usepackage{tcolorbox} %
\usepackage[accsupp]{axessibility}  %

\newcommand{\todo}[1]{{#1}} %

\definecolor{lightgray}{rgb}{0.83, 0.83, 0.83}
\definecolor{Gray}{gray}{0.6}
\definecolor{aliceblue}{rgb}{0.94, 0.97, 1.0}
\definecolor{mistyrose}{rgb}{1.0, 0.89, 0.88}
\definecolor{backcolour}{rgb}{0.95,0.95,0.92}

\newcommand{\fsign}{\text{$V_{1:F}$}}

\newcommand{\tgloss}{\text{$P_{1:G}$}}

\newcommand{\testManual}{\textsc{Sent-Test}\xspace}
\newcommand{\valManual}{\textsc{Sent-Val}\xspace}

\newcommand{\GTPrev}{Prev$^{\text{GT}}$\xspace}
\newcommand{\PredPrev}{Prev$^{\text{Pred}}$\xspace}

\newcommand{\newpara}[1]{\vspace{3pt}\noindent\textbf{#1}}

\usepackage[subtle]{savetrees} %

\newcommand{\appendixref}[2]{%
  \if\sepappendix1%
    #1%
  \else%
    #2%
  \fi%
}

\definecolor{cvprblue}{rgb}{0.21,0.49,0.74}
\usepackage[pagebackref,breaklinks,colorlinks,allcolors=cvprblue]{hyperref}

%% file: authors.tex
\author{
Youngjoon Jang$^{*1,2}$ \hspace{0.4cm}
Haran Raajesh$^{*3,4}$ \hspace{0.4cm}
Liliane Momeni$^{1}$ \hspace{0.4cm} 
G\"{u}l Varol$^{1,3}$ \hspace{0.4cm} 
Andrew Zisserman$^{1}$
\vspace{0.5mm}
\and
$^1$Visual Geometry Group, University of Oxford, UK \\
$^2$KAIST, Daejeon, Republic of Korea \\
$^3$LIGM, École des Ponts, IP Paris, Univ Gustave Eiffel, CNRS, France \\
$^4$CVIT, IIIT Hyderabad, India \\
{\small
\url{https://www.robots.ox.ac.uk/~vgg/research/litfic/}} \\
{\small
$^*$ denotes equal contribution }
}

%% file: 0_abstract.tex
\begin{abstract}

Our objective is to translate continuous sign language into spoken language text.
Inspired by the way human interpreters rely on context for accurate translation, we incorporate additional contextual cues together with the signing video, into a new translation framework. %
Specifically, besides visual sign recognition features that encode the input video, we integrate complementary textual information from
(i)~captions describing the background show,
(ii)~translation of previous sentences, as well as
(iii)~pseudo-glosses transcribing the signing.
These are automatically extracted and inputted along with the visual features to a pre-trained large language model (LLM), which we fine-tune to generate spoken language translations in text form.
Through extensive ablation studies, we show the positive contribution of each input cue to the translation performance. 
We train and evaluate our approach on BOBSL -- the largest British Sign Language dataset currently available. We show that our contextual approach {\em significantly} enhances the quality of the translations compared to previously reported results on BOBSL, and also to state-of-the-art methods that we implement as baselines. Furthermore, we demonstrate the
generality of our approach by applying it also to How2Sign, an American Sign Language dataset, and achieve competitive results. %

\end{abstract}

%% file: 1_intro.tex
\section{Introduction}
\label{sec:intro}
Sign languages are the natural means of communication for deaf communities~\cite{sutton1999linguistics}. They are visual-spatial languages and lack standardised written forms~\cite{filhol:20009:sign-lang:lrec, antinoropizzuto:hal-00665309}. Sign languages exist independently of spoken languages, with their own unique lexicons, ordering, and grammatical structures.
Furthermore, sign languages are expressed through both manual and non-manual components, e.g.\ mouthings, 
which can be conveyed simultaneously~\cite{wilbur2000,Crasborn2006}. 

\begin{figure}
    \centering
    \includegraphics[width=1\linewidth]{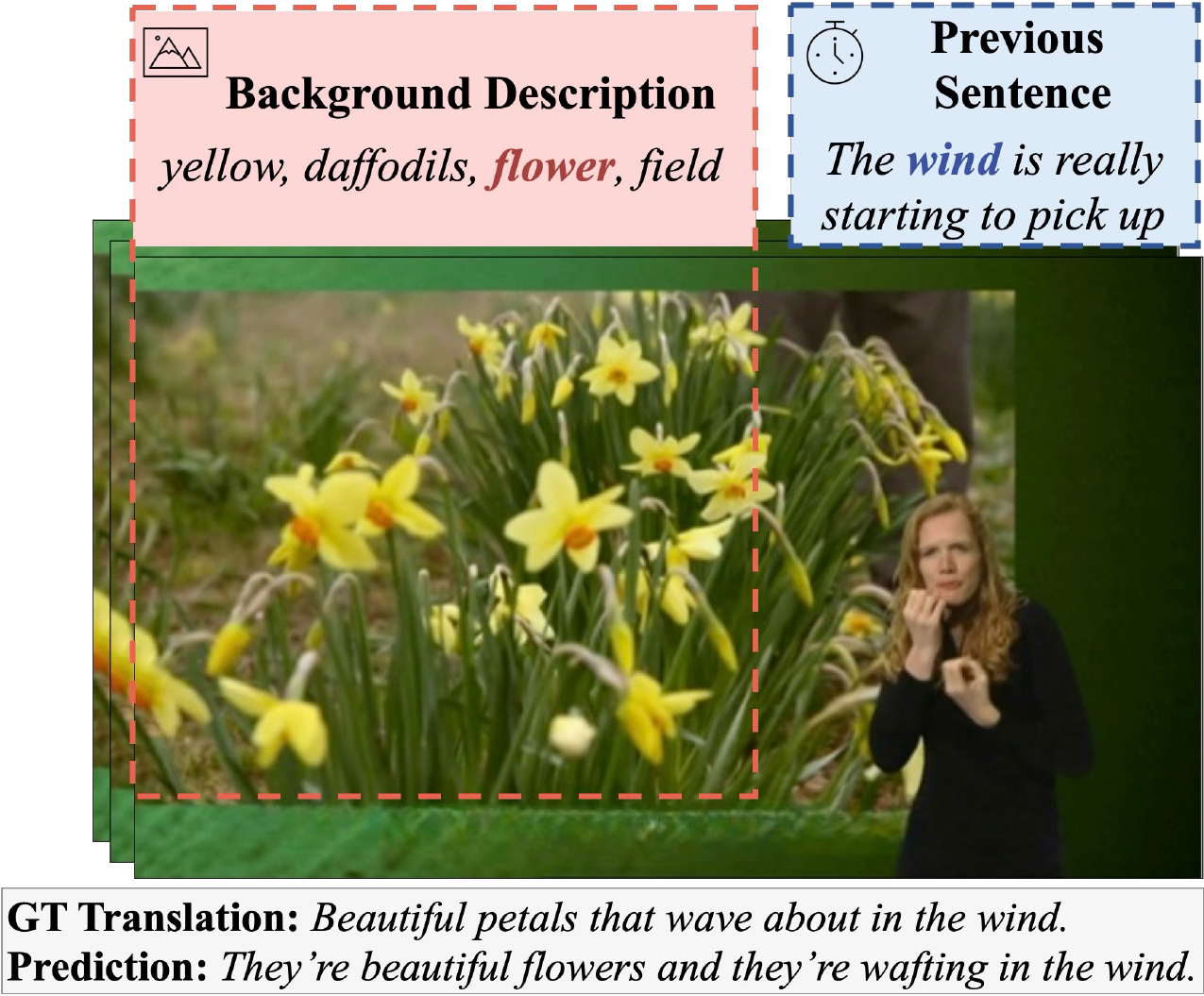}
    \caption{
        \textbf{Contextual cues in SLT}:
        In addition to information extracted from the signing content
        (at the bottom right corner), we give the sign language translation model
        two contextual cues: the background description
        that identifies keywords
        describing the scene behind the signer,
        and the previous sentence translations.
        In this example, the ground truth (GT) translation
        has common words or semantics with the background context (e.g., \textit{flower}),
        and the previous sentence (e.g., \textit{wind}).
    }
    \label{fig:teaser}
\end{figure}

The focus of this work is sign language translation (SLT), the process of transforming sign language into spoken language, in the open-vocabulary setting. Achieving SLT could significantly improve accessibility and inclusion for the deaf and hard-of-hearing communities by reducing communication barriers.

A key challenge in SLT is the fact that sign languages rely heavily on contextual discourse and spatial awareness due to their visual-spatial nature~\cite{muller-etal-2023-findings,reconsideringSSLT2024, Meier_Cormier_Quinto-Pozos_2002,Liddel1990}. Indeed, a study by~\cite{reconsideringSSLT2024} with fluent deaf signers found that a third of their sentence-level signing clips could only be fully translated when provided with additional discourse-level context. 
We next describe three examples of contextual dependencies in sign language:
(a) Signers use spatial indexing to identify referents introduced earlier in the discourse (e.g.\ pointing to identify a particular person, object or any previously defined concept placed in signing space)~\cite{Liddell2003,Emmorey1996,Bergam_1995}; 
(b)~Sign languages typically follow a topic-comment structure~\cite{sutton1999linguistics}. For instance, the topic can refer to setting the temporal framework (e.g.\ if `yesterday' is signed, all verbs in the translation should be in the past tense until a new time is mentioned). 
Once a topic is setup, it is usually not mentioned at every sentence but only re-established when changed, making a sentence-level translation often ill-defined;
(c) Finally, sign language exhibits homonyms~\cite{sutton1999linguistics}, where two signs with similar hand movements can have different meanings %
(e.g.\ `battery' and `uncle' in British Sign Language (BSL)).

Another major obstacle to automatic SLT is the scarcity of large-scale training data. Sign languages are low resource languages, with limited availability of signing videos online, and manually translating signing content is extremely time-consuming, requiring expert annotators. 

In this paper, 
we turn to the underexplored and large-scale BOBSL dataset \cite{Albanie2021bobsl}, which consists of over 1,400 hours of BSL interpreted TV broadcasts with accompanying English subtitles, as our source of training data. 
In comparison to datasets with sentence-level translations~\cite{camgoz-slt,ham2021ksl,zhou2021improving}, interpreters in BOBSL translate spoken language subtitles to signing using context, both from the video playing in the background as well as the previous discourse that has been signed (see \cref{fig:teaser}). We leverage this setting to explore the benefit of context for SLT performance. Specifically, we use \textit{background descriptions} from a captioning model and predicted translations of \textit{previous sentences}, along with sign-level pseudo-glosses\footnote{We %
abuse the linguistic \textit{gloss} term and
refer to sign-level translations in free-form English
as glosses.}
and strong signing visual features as inputs to a pre-trained LLM, which we fine-tune to generate spoken language translations in text form. As signers may use pointing to identify an object or person on the screen, the background descriptions can help with identifying the corresponding referents (e.g.\ there are 4,179 pointing occurrences in the BOBSL-CSLR test set annotations~\cite{raude2024}, which spans only 6 hours). Furthermore, ambiguities of homonyms and tenses/co-references due to a topic-comment structure may be resolved thanks to the context provided from previous sentences,
as well as the background.

However, this task remains challenging due to the \textit{weak} and \textit{noisy} nature of the TV subtitle supervision. The supervision is weak because the subtitles are temporally aligned with the speech, and not perfectly %
with the signing. Although we employ existing automatic signing-subtitle alignment methods~\cite{Bull21}, subtitles may appear a few seconds before or after the corresponding signing sequence. Additionally, the supervision is noisy because words in the subtitle are not necessarily signed, and vice versa. Indeed, sign language interpretation corresponds to a translation -- as opposed to a transcription -- of the speech content, and can often lead to simplification in vocabulary~\cite{bragg2019}. Furthermore, during TV broadcasts, where time pressure is a factor, interpreters may occasionally omit content to keep up with the spoken audio.

In this work, we make the following %
contributions: (i)~we propose a new %
LLM-based model that integrates visual signing and text features with contextual information, including video background descriptions and previous sentence translations; (ii)~we conduct a thorough analysis to examine the impact of each input cue on the translation quality, and %
introduce an LLM-based scoring mechanism that provides a more nuanced translation assessment than traditional metrics such as BLEU~\cite{bleu}; (iii)~we evaluate previous state-of-the-art %
models~\cite{wong2024sign2gpt,zhou2023gloss} on the BOBSL dataset to establish a performance benchmark and find our proposed model surpasses them {\em significantly}; (iv)~we show our contextual method generalises to How2Sign, an American Sign Language (ASL) dataset, demonstrating its effectiveness.

%% file: 2_related.tex
\section{Related Work}
\label{sec:related}

We %
discuss relevant works on isolated and continuous sign language recognition, as well as sign language translation.

\noindent\textbf{Isolated Sign Language Recognition (ISLR)}, which involves
classifying a short single-sign video clip into a sign category,
has been extensively researched over many years. 
In the past decade, ISLR has made significant strides, largely due to the emergence of deep spatiotemporal neural networks and the availability of larger-scale datasets~\cite{Albanie20,AUTSL,Camgz2016BosphorusSignAT,camgoz2016:3dconv,Huang15,Li19wlasl,li2020transferring,Joze19msasl}. In particular, the I3D model~\cite{Carreira2017} has demonstrated its effectiveness in providing robust features for recognition~\cite{Joze19msasl,Li19wlasl,Albanie20}. More recently, the Video Swin Transformer~\cite{Liu2022VideoST} has shown strong ISLR results, and been employed as a backbone for various sign language tasks involving continuous video streams~\cite{Prajwal22a,raude2024}. In this work, we leverage the sign video encoder from~\cite{raude2024} both to obtain visual features and pseudo-glosses.

\noindent\textbf{Continuous Sign Language Recognition (CSLR)} typically involves recognising gloss sequences -- spoken language words corresponding to semantic labels for individual signs -- from a continuous signing video clip.  It is crucial to distinguish CSLR from translation due to the significant grammatical differences between signed and spoken languages~\cite{sutton1999linguistics}. 
Benchmarks such as PHOENIX14~\cite{Phoenix} and CSL-Daily~\cite{zhou2021improving}, which have manually glossed training data, are widely used for CSLR.
Due to a lack of temporal alignment between signs and video frames, many studies~\cite{Cui2019, cheng2020, Wei2023, CoSign, CamgozNecatiCihan2017SEHS, zuo2022c2slr, Hao2021, Pu2019IterativeAN, Min_2021_ICCV, zheng2023cvt} utilise the Connectionist Temporal Classification (CTC) loss~\cite{ctc}. More recently,~\cite{raude2024} demonstrates the advantages of using video-to-text retrieval on the challenging BOBSL dataset \cite{Albanie2021bobsl}.

\begin{figure*}
	\centering
	\includegraphics[width=1.0\linewidth]{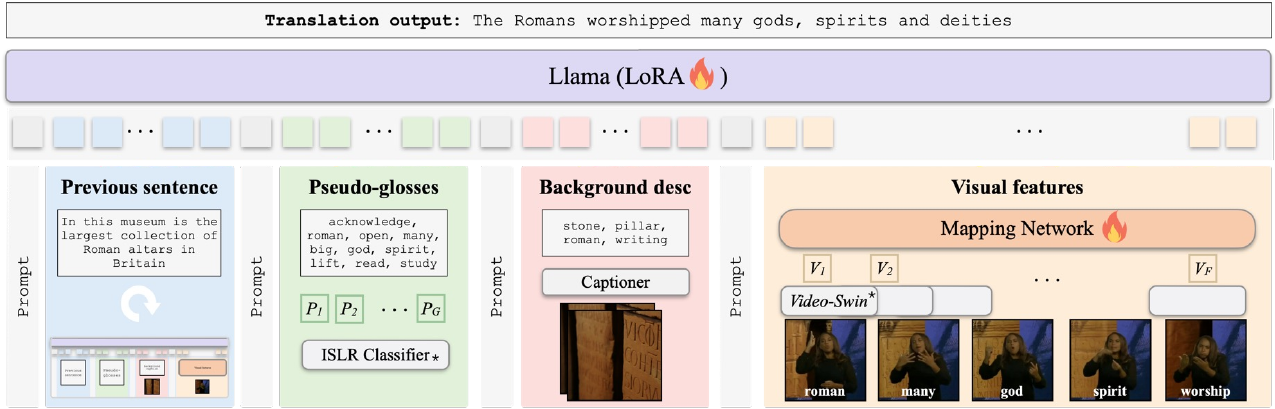}
	\vspace{-0.4cm}
	\caption{\textbf{Method overview:} 
		The input prompt combines contextual cues, the background descriptions and previous sentences, with the information from the current video sequence, specifically visual features and pseudo-glosses. Visual features corresponding to the signer $\{V\}$ are extracted using a pre-trained Video-Swin model,
		which are projected to text space with a learnable mapping network.
		We obtain pseudo-glosses $\{P\}$ by passing
		the Video-Swin features through the pre-trained ISLR Classifier
		(*Video-Swin here denotes the layers except the last one of the ISLR model).
		The background captions, obtained from an off-the-shelf image captioner,
		are summarised into a list of keywords, which we refer to as background descriptions.
		During training, we randomly sample previous GT sentences and previous predictions, while during inference, the model uses its previous prediction in an auto-regressive manner.
		In practice, we include prompts that instruct the model on sign language translation and describe each input.
		We supervise the predicted translation output by comparing it against the ground truth, e.g., 
		`As pagans, the Romans worshipped many gods and spirits'
		in this example.
		Note that we do not use ground-truth glosses -- they are displayed on the bottom right (e.g.\ \textit{roman}, \textit{many}...) only for illustration.
	}
	\label{fig:main}
	\vspace{-0.1cm}
\end{figure*}

\noindent\textbf{Automatic Sign Language Translation (SLT)} involves converting sign language videos into spoken language.
Given that glosses act as a mid-level representation that bridge the visual signing and spoken language modalities, CSLR has been employed as an intermediate step for sign language translation (known as \textit{Sign2Gloss2Text}), or for pre/joint training to enhance visual representations~\cite{camgoz-slt, camgoz2020sign, chen2022two, zhou2021improving, chen2022simple, yin-read-2020-better, ye-etal-2023-cross, zhang2023sltunet, stmc-2021, 10376522}.  
In our work, we only use sign-level pseudo-glosses as an additional input -- rather than as auxiliary supervision -- for translation.

Given the restricted scalability of manual glosses, recent research explores the \textit{gloss-free} SLT setting~\cite{WMT23SLT_knowcomp, muller-etal-2023-findings}, utilising larger datasets and enhanced pre-training techniques. 
For instance, \cite{Albanie2021bobsl} and \cite{slt-how2sign-wicv2023} train SLT systems from scratch using large-scale signing data and robust visual features from a pre-trained ISLR I3D backbone. \cite{shi-etal-2022-open} employs a similar approach, but further pre-trains the visual encoder for sign spotting. \cite{lin-etal-2023-gloss} trains an SLT system jointly with the visual backbone, leveraging conceptual anchor words. GFSLT~\cite{zhou2023gloss} pre-trains their visual encoder and text decoder through a CLIP-style contrastive loss~\cite{clip2021} and a masked self-supervised loss. \cite{ye2024improving} introduces a frame-wise contrastive loss during visual pre-training to enhance feature discrimination. VAP~\cite{jiao2024visual} pre-trains for more fine-grained visual and textual token alignment. \cite{zhang2024scaling} collects large-scale noisy multilingual YouTube SL data and jointly pre-trains for various tasks such as SLT, subtitle-signing alignment, and text-to-text translation. Similarly to these works, we turn to large-scale data and extract strong visual features from an ISLR model.

Recent works also incorporate large-scale language foundation models. For example,~\cite{sincan2024using} inputs pseudo-glosses directly into ChatGPT.
\todo{SignLLM~\cite{gong2024llms} transforms sign language videos into discrete tokens, which are then fed into a frozen language model (LLaMA-7B-16bit~\cite{touvron2023llama}).}
\cite{uthus2023youtubeasl,sandoval-castaneda-etal-2023-ttics,rust-etal-2024-towards,jiao2024visual,zhang2024scaling} all fine-tune a pre-trained T5 model for SLT. \cite{uthus2023youtubeasl} feeds in 3D landmark embeddings, while~\cite{sandoval-castaneda-etal-2023-ttics} and~\cite{rust-etal-2024-towards} use visual features from a BEVT-pre-trained~\cite{bevt} ISLR model and a MAE-pretained video encoder, respectively.  
Sign2GPT~\cite{wong2024sign2gpt} leverages large-scale pre-trained vision (DINOv2~\cite{oquab2023dinov2}) and language (XGLM~\cite{xgml}) models, incorporating adapters (LoRA~\cite{hu2022lora}) for transfer to sign language. Similarly to these works, we make use of a pre-trained LLM (Llama3~\cite{dubey2024llama}) and fine-tune it for SLT.

The most closely related to our work is~\cite{sincan2023context}, that also makes use of context. Specifically, their approach encodes the {\em ground truth} previous subtitle, and spottings (automatic localised sign-level annotations obtained by querying words from the {\em ground truth} subtitle), as well as the signing video before passing them to a transformer decoder to generate translations. Our work differs on multiple fronts. Firstly, our method is fully automatic, employing the \textit{predicted} translation of the previous sentence and \textit{pseudo}-glosses from an ISLR model without access to any ground truth. Second, we additionally incorporate context from the background video. Lastly, we leverage a pre-trained LLM as opposed to training a language decoder from scratch.

%% file: 3_method.tex
\section{Sign Language Translation with Context}
\label{sec:method}

We begin this section by outlining our framework for SLT
using multifarious cues (\cref{subsec:overview}). Next, we describe the representation of each of the
inputs
including visual features, pseudo-glosses, the background description, and the previous sentence (\cref{subsec:representation}). We then present the training strategy that leverages
a pre-trained LLM on this set of inputs 
(\cref{subsec:training}).

\subsection{Framework overview}
\label{subsec:overview}
We introduce a new 
framework to perform SLT in an open-vocabulary setting, basing our model
on a pre-trained LLM.
As illustrated in~\cref{fig:main}, the proposed framework takes various cues as input:
(i)~visual features representing the signing video,
(ii)~pseudo-glosses as a (noisy) automatic transcription of signs,
(iii)~background description as a contextual cue from the TV show displayed behind the signer,
and (iv)~predicted translation of the previous sentence as another contextual cue, obtained auto-regressively at inference.
All modalities except the visual features are in text form, which allows us
to leverage the powerful
language generation capabilities of an LLM, trained on diverse text corpora.
To map the visual features into the space of the LLM input tokens,
we train a simple MLP-based mapping network.
Besides the above inputs, we provide the LLM with an instruction describing the SLT task: ``You are an AI assistant designed to interpret a video of a sign language signing sequence and translate it into English.''
Finally, we add single-sentence prompts in between the inputs by describing each input type, while serving as a separator
(see \appendixref{Appendix~A.2}{\cref{subsec:app:prompts}} for the exact prompts).
All the inputs are appended into a single sequence and fed to the LLM.

\subsection{Model inputs}
\label{subsec:representation}
In the following, we describe how we obtain each of the inputs.

\newpara{Visual features.}
Following \cite{raude2024,Prajwal22a}, we employ
a Video-Swin model~\cite{Liu2022VideoST} pre-trained for ISLR (classification of isolated signs)
to obtain our visual features. Specifically, we utilise the recent and strong model
provided by~\cite{raude2024}, which is trained on the BOBSL videos~\cite{Albanie2021bobsl}
for a vocabulary of 8,697 signs using the automatic sign spottings from~\cite{Momeni22}.
This ISLR model processes short video clips of 16 frames (i.e.\ less than 1 second in 25 fps videos) to produce a single 768-dimensional 
embedding vector. Specifically, this vector comes from spatio-temporal averaging of the penultimate layer output of Video-Swin. 
To capture fine-grained temporal details, we feed 16 consecutive frames into the sign video encoder and apply a sliding window with a stride of $s$ to obtain features \fsign,
where $F$ represents the number of visual features per sentence.
For example, when $s$=2, we have on average 56 features (i.e.\ 4.5 seconds) in BOBSL sentences. 
Note that for experiments conducted on the How2Sign dataset~\cite{Duarte_CVPR2021}, we further fine-tune the Video-Swin model using the spotting annotations provided by~\cite{duarte22slretrieval} for a vocabulary of 1,887 signs (see \appendixref{Appendix~A.5}{\cref{subsec:app:H2S_training}}
for further details).

\newpara{Pseudo-glosses.}
To represent sign sequences in text form, we apply the ISLR model mentioned above
in a sliding window manner
to record the classification predictions,
and obtain $G$ pseudo-glosses \tgloss, representing a sequence of words (or phrases)
from the vocabulary of the classifier (e.g.\ 8k signs). %
Note that these sign category predictions are noisy, often including more
labels than the number of signs occurring in the sentence video, and many false positives,
which we wish to suppress via our LLM tuning.
\todo{For example, \cref{fig:main} shows an example of a  homonym confusion between the manually similar signs `study' and `worship'.}
Our pseudo-glosses are similar to \cite{raude2024},
except we only apply repetition grouping, but do not filter out low-confidence
annotations with a threshold -- this allows the LLM to learn which ones are relevant.
In contrast to \cite{raude2024} that uses pseudo-glosses for supervision,
we simply employ them as additional inputs to our SLT model.
Typically, there are around 22 glosses per BOBSL sentence (which is less than the number of visual features -- 56 features on average).

\newpara{Background description.}
To incorporate context from the background footage,
we crop out the signer from the full frame and apply
an image captioning model (BLIP2~\cite{li2023blip2}) to extract
textual descriptions of the scene behind the signer.
Due to the repetitive nature of captions across consecutive frames
and to reduce computational complexity,
we extract captions at frames sampled at 1 fps, leaving us with 5 captions per signing sentence on average. 
Since similar scenes may persist even over several seconds, resulting in nearly identical captions, we collect all captions per sentence
and keep only the list of unique words (e.g.\ 14 words per signing sentence).
We further remove stopwords (as defined by~\cite{bird-loper-2004-nltk}) to primarily feed keywords that may provide context, and consequently help disambiguate similar signs or identify pointings to the background screen. This process is illustrated in \appendixref{Appendix~A.3}{\cref{subsec:app:background}}.

\newpara{Previous sentence.}
We incorporate the previous sentence text as an additional contextual cue. This refers to the sentence 
that the signer signed leading up to the current one. During training, we use either (i) the ground truth previous sentence or (ii) predictions
precomputed from a model trained without the previous sentence cue (i.e.\ only visual features, pseudo-glosses, and background descriptions).
At test time the model uses its own previous predictions as the previous sentence in an auto-regressive manner.

\subsection{Tuning the LLM with multifarious cues}
\label{subsec:training}
Given the inputs described above, we design and train an LLM-based
model presented in the following.

\newpara{Mapping network for visual features.}
All our inputs are already in text form except the visual features,
which need a projection to map them into the text space, so that they can be fed into the pre-trained LLM.
To this end, we train a simple mapping network,
a 2-layer MLP with GELU~\cite{hendrycks2016gaussian} activation in between,
projecting the visual features (of dimensionality 768) to
the dimensionality of the LLM input embeddings (i.e.\ 4,096).
We note that we add 1D temporal convolution layer to the mapping network when experimenting with the How2Sign dataset \cite{Duarte_CVPR2021}
(see \appendixref{Appendix~A.5}{\cref{subsec:app:H2S_training}}).

\newpara{Training.}
The trainable parameters of our framework are in the mapping network and in the LLM.
We randomly initialise the weights of the mapping network.
For the LLM,
we employ the open-source Llama3 model~\cite{dubey2024llama}, 
specifically opting for the Llama3-8B variant to balance performance and efficiency.
We tune the pre-trained LLM weights to adapt to our SLT task and to our input structure.
Specifically, we adopt LoRA~\cite{hu2022lora} fine-tuning (similarly to~\cite{wong2024sign2gpt}) both to maintain computational efficiency, and not to degrade the powerful language decoding capability of the original model.
We employ the standard cross-entropy loss
across the original LLM vocabulary of 128k text tokens,
using masked self-attention to predict the next tokens.
At inference, the model auto-regressively decodes a sentence until an end-of-sentence token is reached.

\newpara{Augmentations for textual modalities.}
During training,
we apply several augmentations to our textual inputs to enhance model robustness.
First, we perform word dropping on %
three textual cues (pseudo-glosses, %
the previous ground-truth sentence, and the background description) by randomly omitting between 0\% and 50\% of the words in each cue. 
Second, %
we randomly omit entire cues during training, each cue with 50\% probability, allowing the model to flexibly handle inference when some modalities are missing,
but also to make the model pay attention to each cue. %
Third, as previously mentioned,
for previous sentences we make use of both ground-truth and predicted translations during training. %
Again, we randomly sample with a 50\% probability between the two options. This strategy not only serves as an augmentation,
but also reduces the domain gap between training and test time (where we do not have access to ground-truth previous sentences).
In practice, we precompute the predicted previous sentences from a variant of our model trained with all cues except the previous sentence.

\newpara{Implementation details.}
We train on 4 H100 GPUs with a batch size of 2 per GPU, utilising the Adam optimizer~\cite{KingmaAdam}. Training is performed in bfloat16 precision, with FlashAttention-2~\cite{dao2023flashattention} adapted to optimise memory usage.
The LLM decoder (Llama3-8B model~\cite{dubey2024llama})
has a dimensionality 4,096 for its text embeddings.
The LLM is fine-tuned using LoRA with a configuration of rank 4, alpha 16, and dropout 0.05.
We fine-tune only the query and value projectors in all attention layers of the LLM.
Since the text embedding layer has already been pre-trained on a large corpus, we freeze it during training. 
The training spans 10 epochs for BOBSL \cite{Albanie2021bobsl} and 15 epochs
for How2Sign \cite{Duarte_CVPR2021} datasets, including a warmup phase for the first 5 epochs with gradient clipping set to 1.0. The learning rate is set to 0.0001.
We use the HuggingFace library for the pre-trained Llama3 models \cite{dubey2024llama}.

%% file: 4_experiments.tex
\section{Experiments}
\label{sec:experiments}

In this section, we first present the datasets and a suite of evaluation protocols used in our experiments (\cref{subsec:eval-protocol}), as well as baseline descriptions (\cref{subsec:baselines}).
Next, we ablate various components of our framework (\cref{subsec:ablations}). We then show our improved translation performance compared to the state of the art on two challenging open-vocabulary benchmarks (\cref{subsec:sota}). Finally, we illustrate qualitative results and discuss limitations (\cref{subsec:qualitative}).
\todo{Further experiments can be found in
\appendixref{Appendix~B}{\cref{sec:app:experiments}}}.

\subsection{Data and evaluation protocols}
\label{subsec:eval-protocol}

\newpara{BOBSL~\cite{Albanie2021bobsl}} comprises 1,500 hours of BSL-interpreted TV broadcast footage across a wide range of genres, accompanied by English subtitle sentences for the audio content. During training, to achieve better signing-sentence alignments, we use automatically signing-aligned sentences from~\cite{Bull21} as described in~\cite{Albanie2021bobsl}. We filter sentences to those lasting 1-20 seconds as in~\cite{raude2024}, resulting in 689k video-sentence training pairs, corresponding to a vocabulary of 86K words. For evaluation, we utilise the existing validation and test splits, \valManual and \testManual from~\cite{Albanie2021bobsl}, where English sentences have been manually aligned temporally to the continuous signing video. \valManual and \testManual consist of 1,973 and 20,870 aligned sentences, respectively, covering vocabularies of 3,528 and 13,641 English words. We report ablation studies on the validation set and present our final model results on the test set.

\newpara{How2Sign~\cite{Duarte_CVPR2021}} comprises 80 hours of ASL instructional videos from 10 different topics, with temporally aligned sentence language translations. There are 31,128
training, 1,741 validation and 2,322 test sentences,
covering vocabularies of 15.7k, %
3.2k, %
and 3.7k %
English words, respectively. We use the validation set to tune hyperparameters,
and report our final model results on the test set.

\newpara{Evaluation metrics.} To evaluate translation performance effectively, we use five standard evaluation metrics: (i)~BLEU-4 (B4)~\cite{bleu}, which corresponds to the geometric mean of the precision scores of 4-grams, multiplied by a brevity penalty; (ii)~BLEURT (B-RT)~\cite{sellam2020bleurt}, which is a trained metric (using a regression model trained on ratings data) that can better capture non-trivial semantic similarities between sentences; (iii)~ROUGE-L (R-L)~\cite{lin-2004-rouge}, which measures the longest common subsequence between the prediction and ground truth sequence; (iv)~CIDEr~\cite{cider}, a captioning metric that captures consensus of the prediction compared to ground truth by calculating the weighted cosine similarity of TD-IDF scores for various n-grams; and finally, (v)~the Intersection over Union (IoU) of prediction and ground truth token sets (using the Penn Treebank tokenizer~\cite{bird-loper-2004-nltk}). 
Similar to \cite{raude2024,Momeni22}, when computing IoU, we lemmatise all the words, and
do not penalise translated words if they are synoyms to those in ground truth.

\newpara{LLM Evaluation.} Besides these standard metrics, we introduce an LLM-based evaluation metric adapted from the CLAIR framework~\cite{chan-etal-2023-clair} to assess sign language translations. We use the publicly available API of \texttt{GPT-4o-mini} from OpenAI~\cite{gpt4},
prompting the model to generate a score from 0 to 5 for each pair of translation and ground truth, where 5 indicates the best match and 0 is the worst, along with a detailed reasoning for the score. To calibrate the LLM, we provide 12 manually annotated in-context examples. We instruct the LLM to focus on key nouns and verbs and give less importance to pronouns. %
Further details are provided in 
\appendixref{Appendix~A.1}{\cref{subsec:app:llmeval}}.

\subsection{Baselines}
\label{subsec:baselines}

Here, we review baselines that we compare against, including a common translation baseline (Albanie~\cite{Albanie2021bobsl}, Sincan~\cite{sincan2023context}), as well as two state-of-the-art SLT methods (GFSLT~\cite{zhou2023gloss}, Sign2GPT~\cite{wong2024sign2gpt}) according to PHOENIX14T~\cite{Phoenix,camgoz-slt} and CSL-Daily~\cite{zhou2021improving}
benchmarks. %

\newpara{Albanie \cite{Albanie2021bobsl} and Sincan \cite{sincan2023context}.}
These approaches build on the original SLT transformer work of Camgoz~et~al.~\cite{camgoz2020sign}, consisting of a standard transformer encoder-decoder architecture trained from scratch on pre-extracted video frame features.
We compare against the first SLT baseline on BOBSL established by Albanie et al.~\cite{Albanie2021bobsl} using this framework. Specifically, they train the I3D model for 2,281 sign vocabulary on BOBSL spottings, and supervise the translation with automatically-aligned English sentences filtered to a vocabulary of 9k common words. We also compare with the baseline of Sincan et al.~\cite{sincan2023context}, which similarly trains an encoder-decoder model and uses the same I3D features.

\newpara{GFSLT~\cite{zhou2023gloss}.} This recent work introduces a pre-training phase featuring two components: (i)~a CLIP-style contrastive loss, which directs the visual encoder to learn language-aligned visual representations, and (ii)~a masked self-supervised loss, which promotes the ability of the text decoder to grasp sentence semantics. In the subsequent stage, the pre-trained visual encoder and text decoder are jointly fine-tuned within an encoder-decoder translation framework, enabling the direct conversion of visual representations into spoken sentences.
We first reproduce this method using the public codebase on PHOENIX14T
(see \appendixref{Appendix~B.7}{\cref{subsec:app:phoenix}}), before adapting
to the BOBSL dataset.

\newpara{Sign2GPT~\cite{wong2024sign2gpt}.} This state-of-the-art work proposes an encoder-decoder translation framework that leverages large-scale pre-trained vision (DINOv2~\cite{oquab2023dinov2}) and language (XGLM~\cite{xgml}) models, incorporating adapters (LoRA~\cite{hu2022lora}) for transfer to sign language. Additionally, a prototype-driven pre-training strategy is introduced, which guides the visual encoder to learn sign representations from spoken language sentences by filtering specific parts of speech.
We again use the public codebase for reproducing and applying on the BOBSL dataset.
We report results both without and with pre-training (denoted as w/PGP) as in~\cite{wong2024sign2gpt}.

\newpara{Oracle: Sincan~\cite{sincan2023context} (Vid+\GTPrev).} This oracle baseline is a multi-modal variant of Sincan~\cite{sincan2023context}, where the %
previous {\em ground truth} sentence is also fed as additional context to the decoder at both training and inference times. 
 
\newpara{Oracle: Sincan~\cite{sincan2023context} (Vid+\GTPrev+Spot).} This baseline is another oracle variant, where \textit{Spottings} are also fed into the decoder at both training and inference times. Spottings are automatic sign annotations~\cite{Momeni22}, obtained using {\em ground truth} knowledge of the %
nearby English subtitles,
i.e.\ given words from the subtitle, by temporally localising them in video. We note that our pseudo-glosses are different to \textit{Spottings} as they are predicted directly from the video, without access to the corresponding English sentence translation.

\subsection{Ablation study}
\label{subsec:ablations}

In this section, we analyse our different design choices. We present our results on the BOBSL validation set, \valManual, using BLEURT (B-RT), IoU and LLM evaluation metrics. 

\newpara{Combining different cues.}
In \cref{tab:combining-cues}, we
measure the contribution of each input cue.
With only visual features, the model achieves baseline scores of 41.0 for B-RT, 16.6 for IoU, and 1.29 for LLM score.
Adding pseudo-glosses (i.e.\ textual cues derived from the current sign video) improves all metrics, highlighting the benefit of text directly related to the signs. Incorporating previous sentences as a contextual cue further boosts performance, and finally, adding background descriptions achieves the best results across all metrics. Overall, the final model, with all cues combined, yields a significant improvement of +2.5 in B-RT, +2.2 in IoU, and +0.27 in LLM score compared to the baseline,
confirming that there is additional relevant information found in context, beyond the signing video, to help with the translation.
In 
\appendixref{Appendix~A.1}{\cref{subsec:app:llmeval}}, we provide further statistics analysing the distribution of the LLM evaluation scores, which align well with other metrics in most cases, while being interpretable.

\begin{table}
	\small
	\setlength{\tabcolsep}{9pt}
	\centering
	\resizebox{0.99\linewidth}{!}
	{
		\begin{tabular}{cccc|lll}
			\toprule
			Vid & PG & \PredPrev  & BG & B-RT & IoU & LLM  \\
			\midrule
			\checkmark &  &   & & 41.0 & 16.6 & 1.29\\
			\checkmark & \checkmark &   & & 41.8 & 17.5 & 1.40\\
			\checkmark & \checkmark & \checkmark  & & 42.5 & 18.1 & 1.45\\
			\rowcolor{aliceblue} \checkmark & \checkmark & \checkmark  &  \checkmark & \textbf{43.5} & \textbf{18.8} & \textbf{1.56} \\
			\bottomrule
		\end{tabular}
	}
	\vspace{-0.2cm}
	\caption{
		\textbf{Combining different cues.} We analyse on BOBSL \valManual, how different cues contribute to translation performance, when added to the vanilla model
		inputting only the visual signing features (\textit{Vid}). We observe that pseudo-glosses (\textit{PG}), background description (\textit{BG}), and predicted translation of the previous sentence (\textit{\PredPrev}) are all complementary, as combining them achieves the best performance.
	}
	\vspace{-0.2cm}
	\label{tab:combining-cues}
\end{table}

\begin{table}
    \small
    \setlength{\tabcolsep}{5pt}
    \centering
    {
        \begin{tabular}{c|ccc|lll}
        \toprule
        LLM &    Drop & Drop & \PredPrev/ & B-RT & IoU & LLM   \\
        fine-tuning & words & cue & \GTPrev & &    \\
        \midrule
        \checkmark &    & & & 41.2 & 17.0 & 1.40 \\
        \checkmark &    \checkmark & & & 41.4 & 17.4 & 1.41\\
        \checkmark &    \checkmark & \checkmark & & 42.7 & 18.1 & 1.53\\
        \midrule
        \rowcolor{aliceblue}   \checkmark &  \checkmark & \checkmark &  \checkmark  & \textbf{43.5} & \textbf{18.8} & \textbf{1.56} \\
        &  \checkmark & \checkmark &  \checkmark  &  40.6 & 16.7 & 1.27 \\
        \bottomrule
        \end{tabular}
    }
    \vspace{-0.2cm}
    \caption{
	\textbf{Augmentations and LLM fine-tuning.} We ablate, on BOBSL \valManual,
	how different input augmentations and fine-tuning the LLM decoder with LoRA~\cite{hu2022lora} impact the %
	performance. %
	As explained in \cref{subsec:training}, we randomly \textit{drop words} within each cue,
	or entirely \textit{drop a cue}.
	\textit{\PredPrev/\GTPrev} refers to randomly sampling either the predicted or ground-truth translation for the previous sentence.
    We observe that the combination of all input augmentations leads to the best %
    performance,
    and also show the benefits of LoRA fine-tuning.
    }
    \vspace{-0.2cm}
    \label{tab:augmentation}
\end{table}

\newpara{Effect of augmentations.} 
We perform a series of ablations in \cref{tab:augmentation} regarding %
the input augmentations at training time: (i)~\textit{Drop words}: randomly removing up to 50\% of the words in textual cues, (ii)~\textit{Drop cue}: randomly removing an entire cue with a probability of 50\%, and (iii)~\textit{\PredPrev/\GTPrev}: sampling either predicted or ground truth (GT) previous sentence with equal probability.
When \textit{Drop words} augmentation is applied, we observe a slight performance increase.
Adding \textit{Drop cue} augmentation provides an additional improvement (42.7 vs 41.4). We hypothesise that these augmentations make the model less sensitive to noise and less reliant on any particular textual cue.
Finally, combining \textit{\PredPrev/\GTPrev} augmentation further boosts translation performance (43.5 vs 42.7), as it reduces the reliance on previous GT sentences and better matches the inference setting.

\newpara{LLM fine-tuning.} 
In \cref{tab:augmentation}, we also examine a model variant without fine-tuning the LLM, but only training the mapping network with a frozen LLM. Comparing the last two rows, we observe that LoRA fine-tuning the decoder yields improvements across all metrics (B-RT: +2.9, IoU: +2.1, and LLM score: +0.29).
We hypothesise that the improvement may partially be due to distinct
linguistic characteristics of signed and spoken languages, but also
due to adapting the LLM to our specific input structure.
\begin{table}
    \small
    \setlength{\tabcolsep}{3pt}
    \centering
    \resizebox{1\linewidth}{!}
    {
    \begin{tabular}{l|cccccc}
        \toprule
        Model  & B4 & B-RT & R-L & CIDEr & IoU & LLM \\
        \bottomrule
        \rowcolor{gray!10} \multicolumn{7}{c}{\textsc{Video only}} \\
        Albanie ~\cite{Albanie2021bobsl} & 1.0 & - & 10.2 & - & - & - \\ 
        Sincan~\cite{sincan2023context}  & 1.3 & - & 8.9 & - & - & -\\
        GFSLT~\cite{zhou2023gloss} $\dagger$ & 0.6 & 27.7 & 7.4 & 4.3 & 5.2 & 0.05 \\ 
        Sign2GPT~\cite{wong2024sign2gpt} $\dagger$ & 0.7 & 34.3 & 10.6 & 12.8 & 8.2 & 0.37 \\ 
        Sign2GPT (w/PGP)~\cite{wong2024sign2gpt} $\dagger$ & 0.9 & 35.2 & 11.4 & 16.1 & 8.7 & 0.49 \\ 
        \rowcolor{aliceblue} \todo{Ours (Vid)} & 2.6 & 37.8 & 15.6 & 37.5 & 13.6 & 0.95 \\ 
        \bottomrule
        \rowcolor{gray!10} \multicolumn{7}{c}{\textsc{Extra cues}} \\

        \rowcolor{aliceblue} \todo{Ours (Vid+\PredPrev{})} & 2.8 & 38.9 & 16.0 & 37.8 & 13.8 & 1.02 \\
        \rowcolor{aliceblue} \todo{Ours (Vid+\PredPrev{}+PG)} & 2.9 & 39.2 & 16.1 & 38.9 & 14.1 & 1.15 \\
        \rowcolor{aliceblue} \todo{Ours (Vid+\PredPrev{}+PG+BG)} & \textbf{3.3} & \textbf{40.3} & \textbf{16.9} & \textbf{41.9} & \textbf{14.8} & \textbf{1.20} \\
        \bottomrule
        \rowcolor{gray!10} \multicolumn{7}{c}{\textsc{Oracle}} \\
        Sincan~\cite{sincan2023context}  (Vid+\GTPrev) & 1.5 & 35.8 & 9.7 & 23.9 & 10.4 & 0.56\\
        \rowcolor{aliceblue} \todo{Ours (Vid+\GTPrev{})}  & 3.3 & 40.5 & 17.3 & 42.8 & 14.8 & 1.23 \\
        \midrule
         Sincan~\cite{sincan2023context}  (Vid+\GTPrev+Spot)  & 2.9 & 37.0 & 12.4 & 41.0 & 12.5 & 0.80 \\
        \rowcolor{aliceblue} \todo{Ours (Vid+\GTPrev{}+Spot)} & 6.5 & 45.9 & 24.0 & 82.5 & 25.5 & 1.69 \\
         \rowcolor{aliceblue} Ours (Vid+\GTPrev{}+Spot+BG) & \textbf{7.3} & \textbf{47.1} & \textbf{25.1} & \textbf{88.9} & \textbf{26.5} & \textbf{1.85} \\
        \bottomrule
    \end{tabular}
    }
    \vspace{-0.2cm}
    \caption{
    \textbf{Comparison to the state of the art on BOBSL \testManual.} We compare our method to previous state-of-the-art works and surpass their performance on a range of translation metrics. In the \textsc{Oracle} setting (bottom block), we compare fairly to approaches which use (i)~the previous ground truth sentence as context (\GTPrev),
    as opposed to the predicted previous sentence
    (\PredPrev), and (ii)~Spottings (Spot) that are derived from the current ground truth sentence, as opposed to sign-level pseudo-glosses (PG). For example, in `Ours (Vid+\GTPrev{}+Spot+BG)' we replace our pseudo-glosses with the spottings
    that have access to ground truth sentence, to show a more similar setting to~\cite{sincan2023context}. $\dagger$ denotes scores that we obtained by training methods of \cite{zhou2023gloss,wong2024sign2gpt} on BOBSL. Note that unlike previous experiments on the validation set, this table reports on the test set.
    }
    \vspace{-0.3cm}
    \label{tab:sota-bobsl}
\end{table}

\subsection{Comparison to the state of the art}\label{subsec:sota}

\newpara{BOBSL.} We evaluate our model on the BOBSL test set, \testManual, using our full suite of evaluation metrics. 
\todo{As shown in \cref{tab:sota-bobsl}, our approach achieves a significant improvement across all metrics compared to previous works. Notably, even with only video input, our model surpasses state-of-the-art methods such as GFSLT~\cite{zhou2023gloss} and Sign2GPT~\cite{wong2024sign2gpt}. Moreover, incorporating additional cues leads to a steady performance gain, with the full set of cues yielding a considerable boost (40.3 vs 37.8 B-RT). This highlights both the effectiveness of our method, which leverages context and the increased challenge posed by the BOBSL dataset (as opposed to PHOENIX14T where \cite{zhou2023gloss,wong2024sign2gpt} were originally evaluated).}

In the \textsc{Oracle} setup (the bottom block of \cref{tab:sota-bobsl}),
we compare to the setting of \cite{sincan2023context},
where models have access to \textit{ground truth} previous sentence and
spottings extracted from the \textit{ground truth} current sentence.
When using the ground truth previous sentence at inference, our model outperforms~\cite{sincan2023context} by a large margin 
\todo{(40.5 vs 35.8 B-RT).}
\todo{When using both the ground truth previous sentence and \textit{spottings} (which are obtained with access to current ground truth),} 
we further increase the margin,
substantially
outperforming their method
\todo{(45.9 vs 37.0 B-RT)}.
\todo{Additionally, when integrating background descriptions (last row), we observe a further performance gain}.

\newpara{How2Sign.}
Here, we demonstrate the generality of our method
by training on the How2Sign dataset (see
\appendixref{Appendix~A.5}{\cref{subsec:app:H2S_training}} for details).
In \cref{tab:sota-how2sign}, we compare against the state of the art
on the test set, and also report variants of our model by gradually adding more cues.
We observe that adding the pseudo-glosses, as well as the contextual cue of the previous translated sentence boosts performance. We note that in this case, we do not use background descriptions since How2Sign does not consist of interpreted TV with an accompanying show. 
We find our best model (Vid+PG+\PredPrev) achieves comparable performance with the state-of-the-art method VAP~\cite{jiao2024visual} in terms of B4 and attains a higher R-L score by nearly 5 points (32.5 vs 27.8). We note that we include the numbers from \cite{zhang2024scaling} and \cite{rust-etal-2024-towards} (denoted with $\dagger$), however, we do not compare to these as they train SLT additionally on a large ASL corpus of 984 hours~\cite{uthus2023youtubeasl}.

\begin{table}
    \small
    \setlength{\tabcolsep}{3pt}
    \centering
    \resizebox{1\linewidth}{!}
    {
        \begin{tabular}{l|cccccc}
        \toprule
        Model  & B4 & B-RT & R-L & CIDEr & IoU & LLM \\
        \midrule
        \textcolor{gray}{SSLT~\cite{zhang2024scaling}} $\dagger$ & - & \textcolor{gray}{55.7} & - & - & - & -\\ 
        \textcolor{gray}{SSVP-SLT~\cite{rust-etal-2024-towards}} $\dagger$ & \textcolor{gray}{15.5} & \textcolor{gray}{49.6} & \textcolor{gray}{38.4} & - & - & - \\ 
        \midrule
        SSLT~\cite{zhang2024scaling} & - & 34.0 & - & - & - & -\\ 
        SSVP-SLT~\cite{rust-etal-2024-towards} & 7.0 & 39.3 & 25.7 & - & - & -\\ 
        Fla-LLM~\cite{chen2024factorized} & 9.7 & - & 27.8 & - & - & - \\ 
        VAP~\cite{jiao2024visual} & \textbf{12.9} & - & 27.8 & - & - & - \\ 
        \midrule
        \rowcolor{aliceblue} Ours (Vid)  & 11.8 & 44.1 & 31.1 & 93.3 & 26.1 & 1.39 \\
        \rowcolor{aliceblue} Ours (Vid+PG)  & 12.3 & 44.7 & 31.9 & 97.8  & 27.4 & 1.55 \\
        \rowcolor{aliceblue} Ours (Vid+PG+\PredPrev)  & 12.7 & \textbf{45.3} & \textbf{32.5} & \textbf{100.8}  & \textbf{27.9}\ & \textbf{1.59} \\
        \bottomrule
        \end{tabular}
    }
    \vspace{-0.2cm}
    \caption{
    \textbf{Comparison to the state of the art on How2Sign.} We compare our method to previous works that report on the How2Sign test set,
    and obtain competitive performance. We also observe advantages of incorporating additional cues from pseudo-glosses (PG) and previous predicted sentence (\PredPrev).
    $\dagger$ denotes methods that pre-train the SLT model on a larger ASL dataset (YouTube-ASL~\cite{uthus2023youtubeasl} which covers 984 hours).
    }
    \vspace{-0.3cm}
    \label{tab:sota-how2sign}
\end{table}

\begin{figure*}
    \centering
    \includegraphics[width=0.49\linewidth]{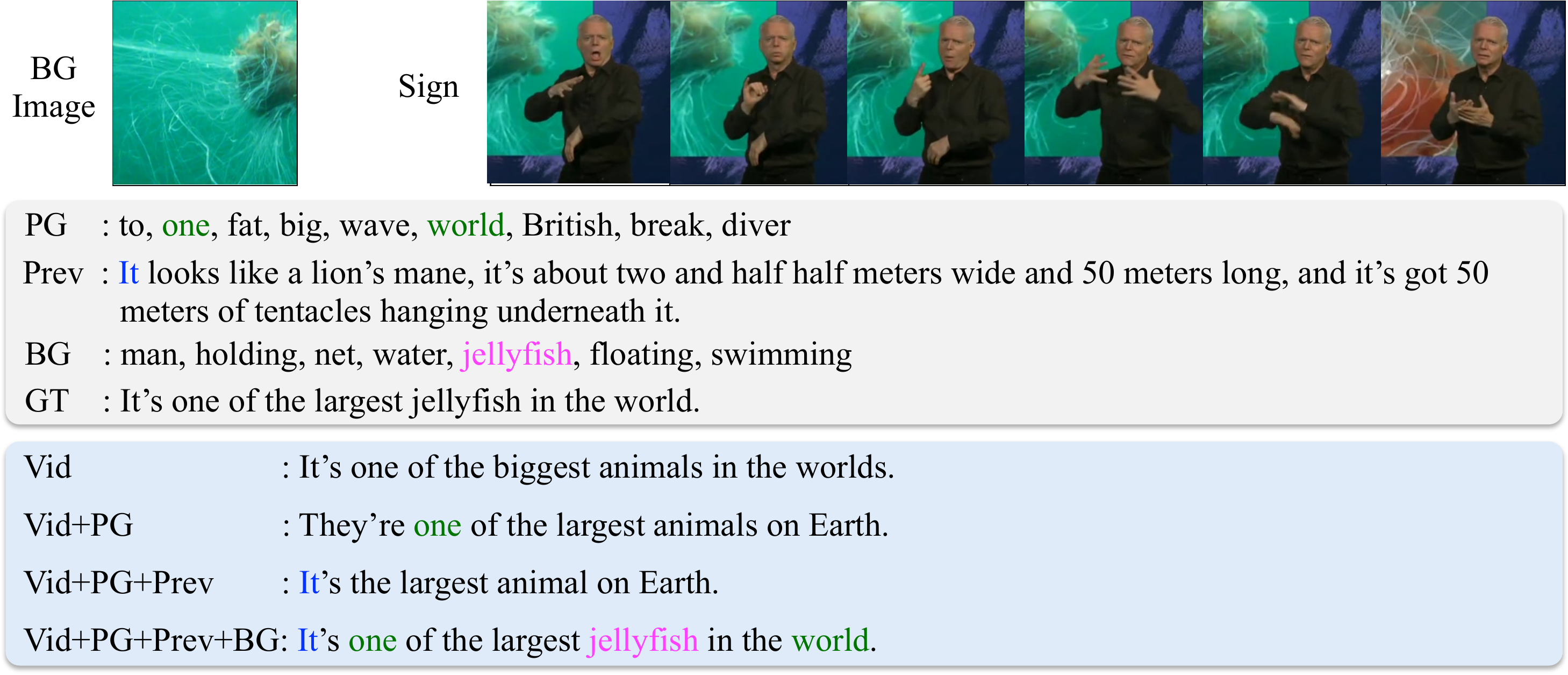}\hfill
    \includegraphics[width=0.49\linewidth]{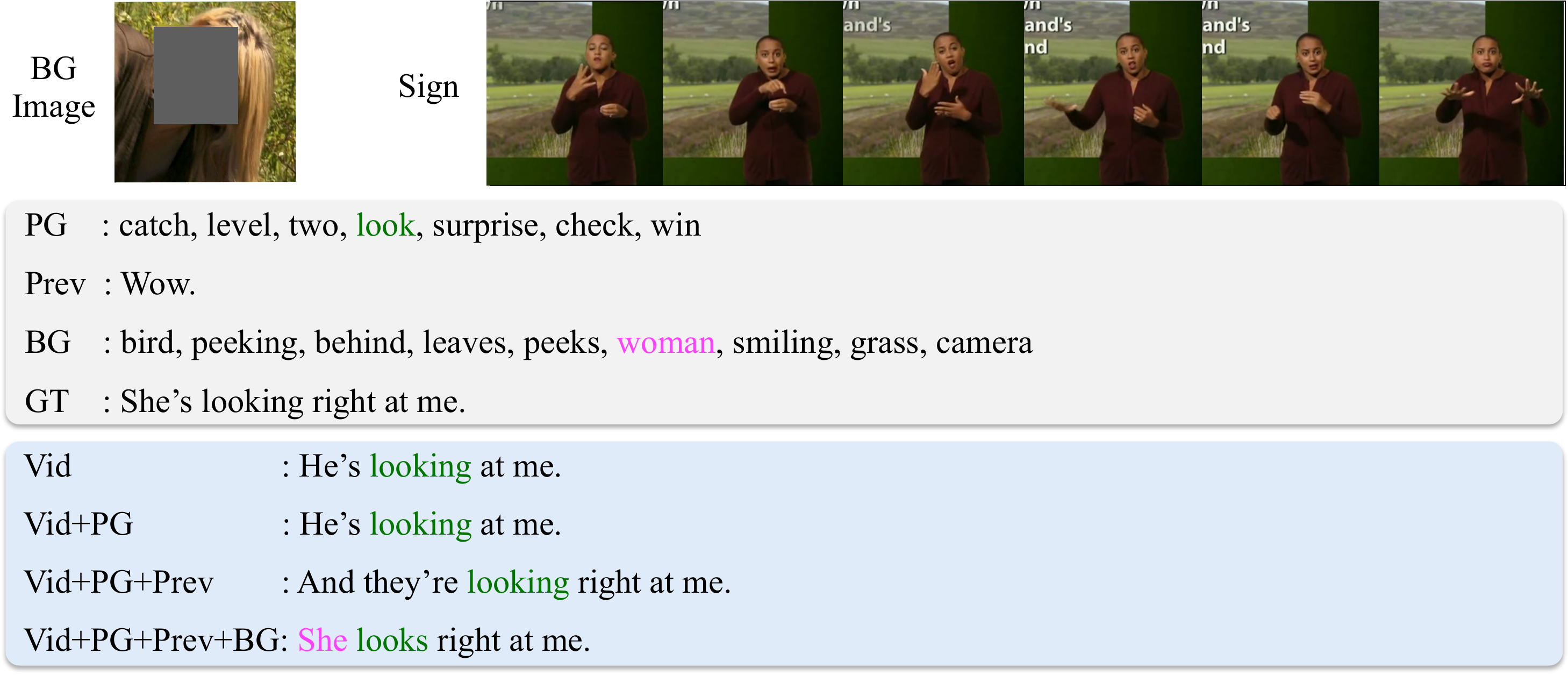}\vspace{0.3cm}
    \includegraphics[width=0.49\linewidth]{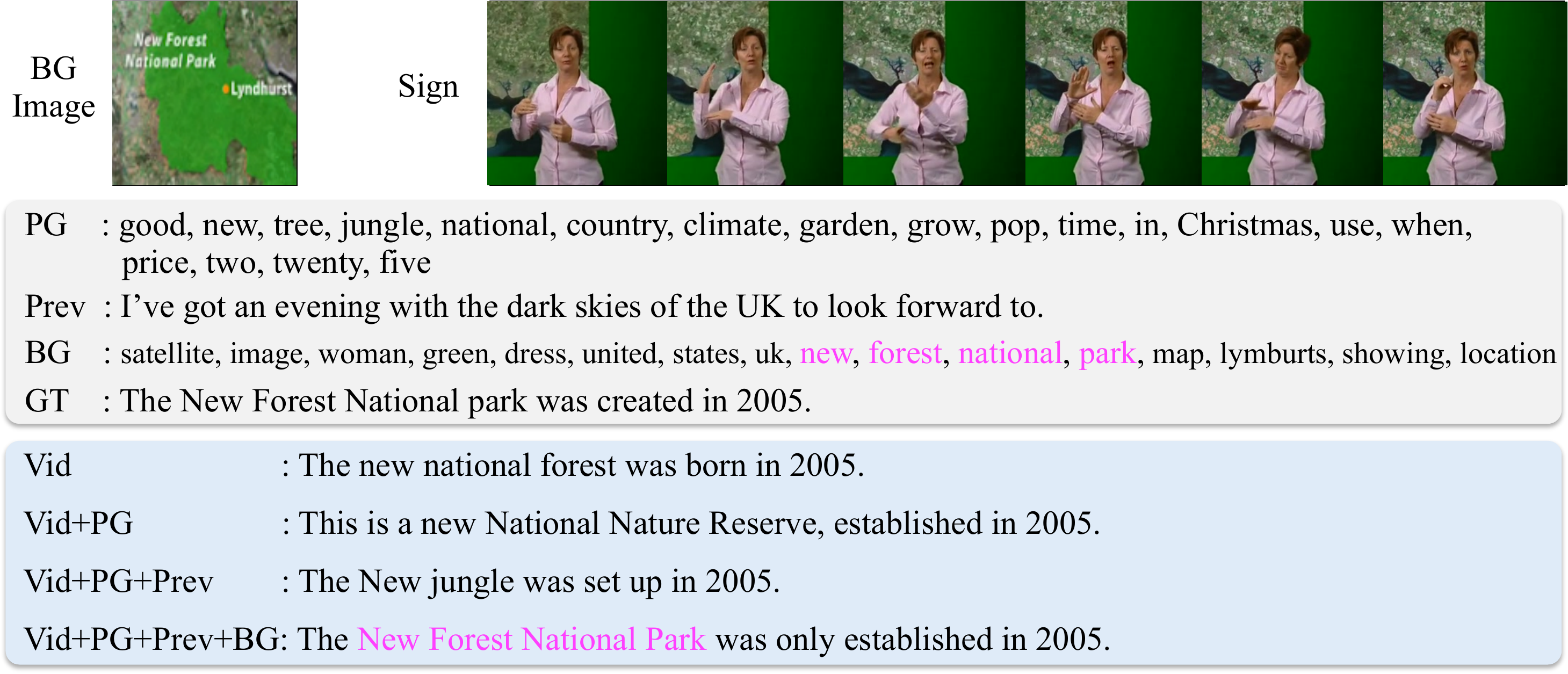}\hfill
    \includegraphics[width=0.49\linewidth]{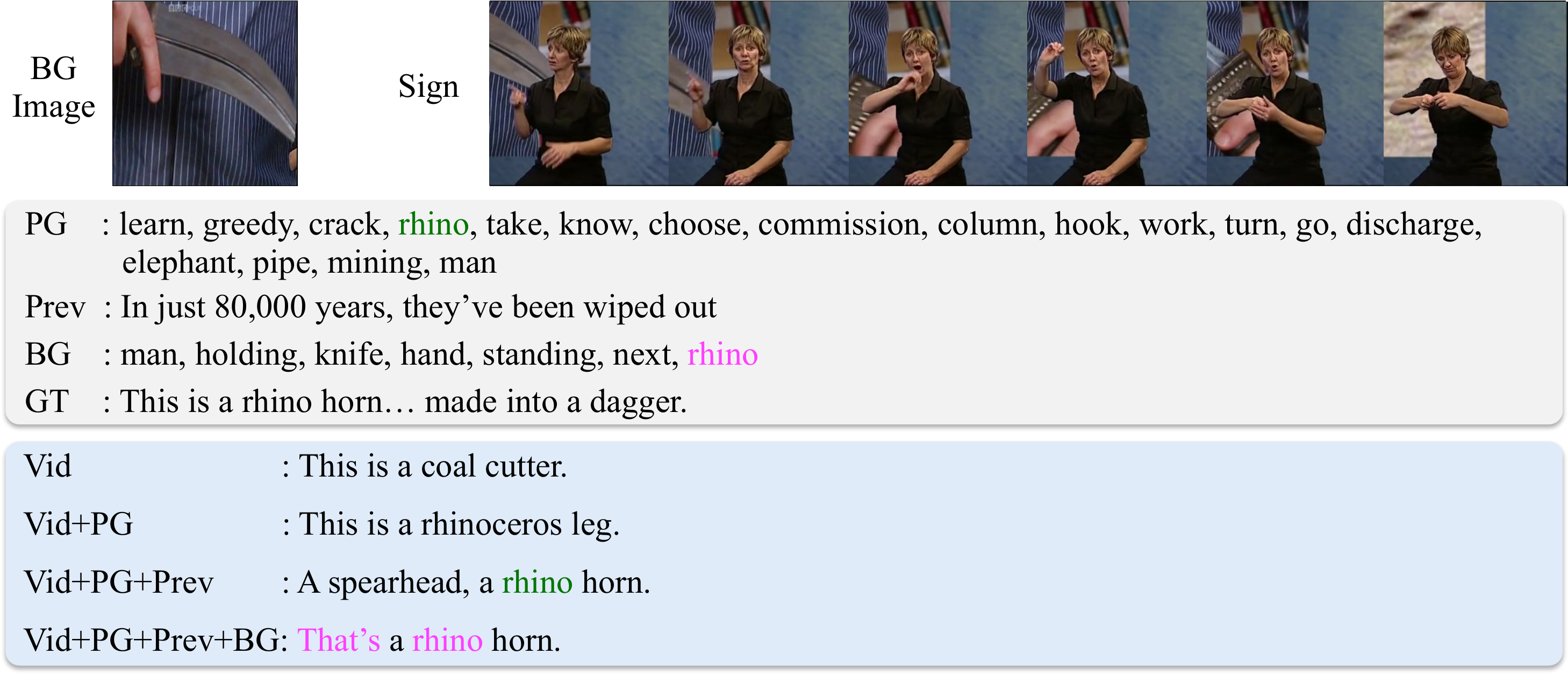}\vspace{0.3cm}
    \includegraphics[width=0.49\linewidth]{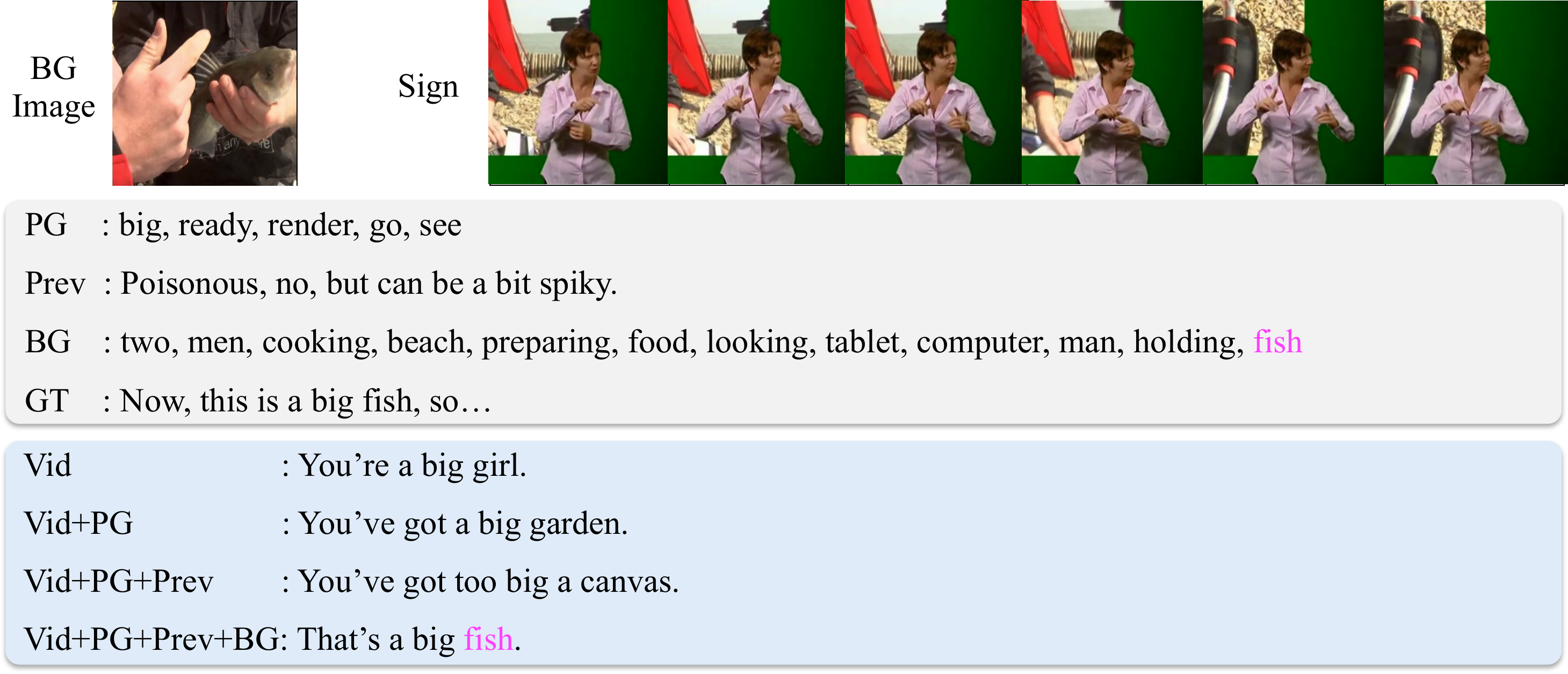}\hfill
    \includegraphics[width=0.49\linewidth]{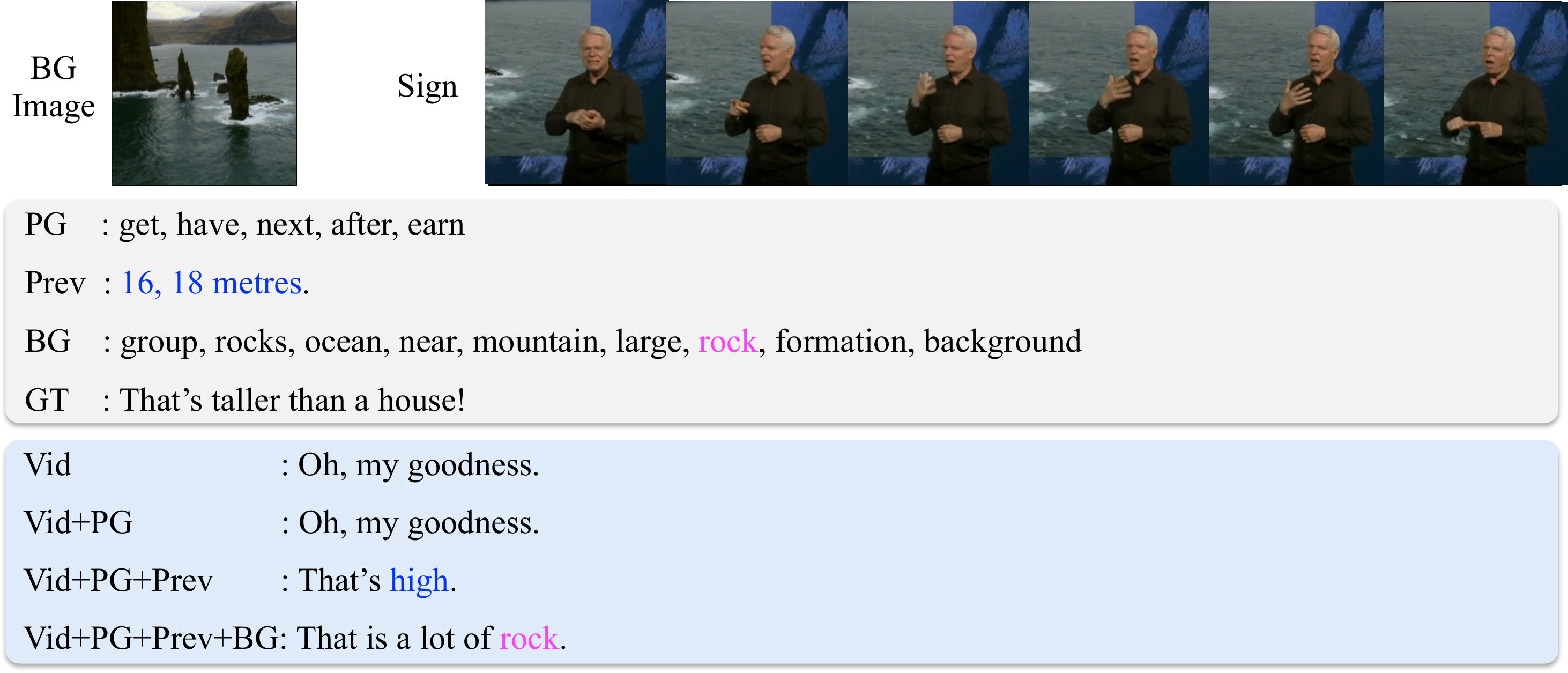}
    \vspace{-0.2cm}
    \caption{
        \textbf{Qualitative analysis:} We present visual examples to show how different cues affect the translation results. Starting with visual features, we incrementally add pseudo-glosses (PG), the predicted previous sentence (Prev), and the background description (BG).
        We observe that the previous sentence helps translation performance by providing further context (top left, bottom right). The background description also helps for pronoun referencing (top right), place names (middle left), pointing gestures (middle right), and object referencing (bottom left). However, the background can also in some cases hinder translation (bottom right). We refer to \cref{subsec:qualitative} for detailed comments.
    }
    \label{fig:qualitative}
    \vspace{-0.2cm}
\end{figure*}

\subsection{Qualitative analysis and limitations}\label{subsec:qualitative}

We visually analyse how different input cues impact the translation outputs by providing relevant information beyond the signing video. In \cref{fig:qualitative} (top left), a key focus of the sentence -- the word \textit{jellyfish} -- is not signed. In \cref{fig:qualitative} (top right), the sign for pronouns \textit{he} and \textit{she} is ambiguous. In such cases, the model needs to utilise available context -- much like a human interpreter would -- to accurately translate the sentence. By effectively leveraging the background context in both cases, the model is able to produce stronger translations. In other cases, context can be used to further augment information obtained from the video. This can be seen in \cref{fig:qualitative} (middle left), where using only the video, the model gains a general theme about \textit{national forest} but by leveraging the context, can precisely generate \textit{New Forest National Park}. \cref{fig:qualitative} (middle right) shows an example where the signer points to the background where the \textit{rhino horn} appears on screen, and the similar sign for \textit{elephant} that appears in pseudo-glosses is effectively suppressed.

However, we do observe several challenges as well: (i)~different cues may present conflicting information, and while the model often learns to implicitly resolve such conflicts, there are cases where it struggles (see \cref{fig:qualitative}, bottom right);
(ii)~our model can struggle to discern the grammatical context of a sentence, e.g.\ it sometimes cannot distinguish whether a given sentence is a question or a statement; 
(iii)~similarly the model makes frequent mistakes by missing negations;
(iv)~our model still faces difficulties with certain sign types, such as pointing and fingerspelling, which are essential components of sign language. These limitations highlight the complexity of sign language translation and underscore the need for continued research and development in the field. Additional qualitative results are provided in \appendixref{Appendix~C}{\cref{sec:app:qualitative}}.

%% file: 5_conclusion.tex
\section{Conclusion}
\label{sec:conclusion}
In this work, we show that leveraging contextual information significantly enhances SLT performance in an open-vocabulary setting. Specifically, our framework utilises background descriptions from a captioning model and predictions of previous sentences, combined with pseudo-glosses and visual features. 
Through an extensive ablation study, we analyse the individual impact of each cue on sign language translation, and benchmark our method against previous state-of-the-art approaches on the BOBSL dataset to demonstrate its effectiveness.
\todo{Despite the improvements of our approach, several challenges remain. Background descriptions may not always complement the translation, potentially introducing noise. Relying on previous sentence predictions can lead to error accumulation, where early mistakes propagate through subsequent sentences. Additionally, pseudo-glosses are prone to false positives due to homonyms, resulting in incorrect word choices. Addressing these limitations would potentially improve the robustness and real-world applicability of sign language translation systems.}

%% file: 6_acknowledgements.tex
{
\paragraph{Acknowledgments.}
The images in this paper are used with kind permission of the BBC.
This work was granted access to the HPC
resources of IDRIS under the allocation 2024-AD011013395 made
by GENCI. The authors would like to acknowledge the ANR project
CorVis ANR-21-CE23-0003-01, a Google Research Scholar Award, and a Royal Society
Research Professorship RP$\backslash$R1$\backslash$191132.
The authors also thank
Charles Raude, Prajwal KR, Prasanna Sridhar, Joon Son Chung, Makarand Tapaswi, Syrine Kalleli, and David Picard 
\todo{for their feedback,
and further thank Ryan Wong for guiding the Sign2GPT~\cite{wong2024sign2gpt} implementation and Ozge Mercanoglu Sincan for providing the outputs for the Sincan~\cite{sincan2023context} model on the
BOBSL test set.
}

%% file: 7_appendix.tex
\appendix

\renewcommand{\thefigure}{A.\arabic{figure}} %
\setcounter{figure}{0} 
\renewcommand{\thetable}{A.\arabic{table}}
\setcounter{table}{0} 

This appendix supplements the main paper by providing additional
implementation details (\cref{sec:app:implementation}),
experiments (\cref{sec:app:experiments}),
and qualitative results (\cref{sec:app:qualitative}).

\startcontents[sections]
{
	\hypersetup{linkcolor=black}
	\printcontents[sections]{l}{1}{}
}

\section{Implementation Details}
\label{sec:app:implementation}
We provide details on the LLM evaluation metric (\cref{subsec:app:llmeval}),
the list of prompts for the translation model (\cref{subsec:app:prompts}),
the background description collation (\cref{subsec:app:background}),
the architectural design for the mapping network on BOBSL (\cref{subsec:app:bobsl_mapping}),
and
the training procedure on How2Sign (\cref{subsec:app:H2S_training}).

\subsection{LLM Evaluation metric}
\label{subsec:app:llmeval}
As mentioned in \appendixref{Sec.~4.1}{\cref{subsec:eval-protocol}}, our LLM-based evaluation metric is adapted from the CLAIR framework~\cite{chan-etal-2023-clair}. 
Here, we detail the prompts and show an analysis for this metric.

\newpara{LLM evaluation prompt.}
\cref{fig:app:llmevalprompt} shows the system, user, and assistant prompts,
that we input to \texttt{GPT-4o-mini}~\cite{gpt4},
to define the sign language translation evaluation task.
To calibrate the language model,
we include 12 manually annotated in-context examples,
displayed in \cref{tab:app:incontext}, with two examples per score from 0 to 5. 
Each example contains both the score and the reasoning according to our instructions, focusing on key nouns and verbs, while giving less importance to pronouns. 
This approach makes the metric interpretable, as the LLM outputs detailed reasoning for each score. 

\begin{figure*}
    \centering
    \includegraphics[width=0.32\linewidth]{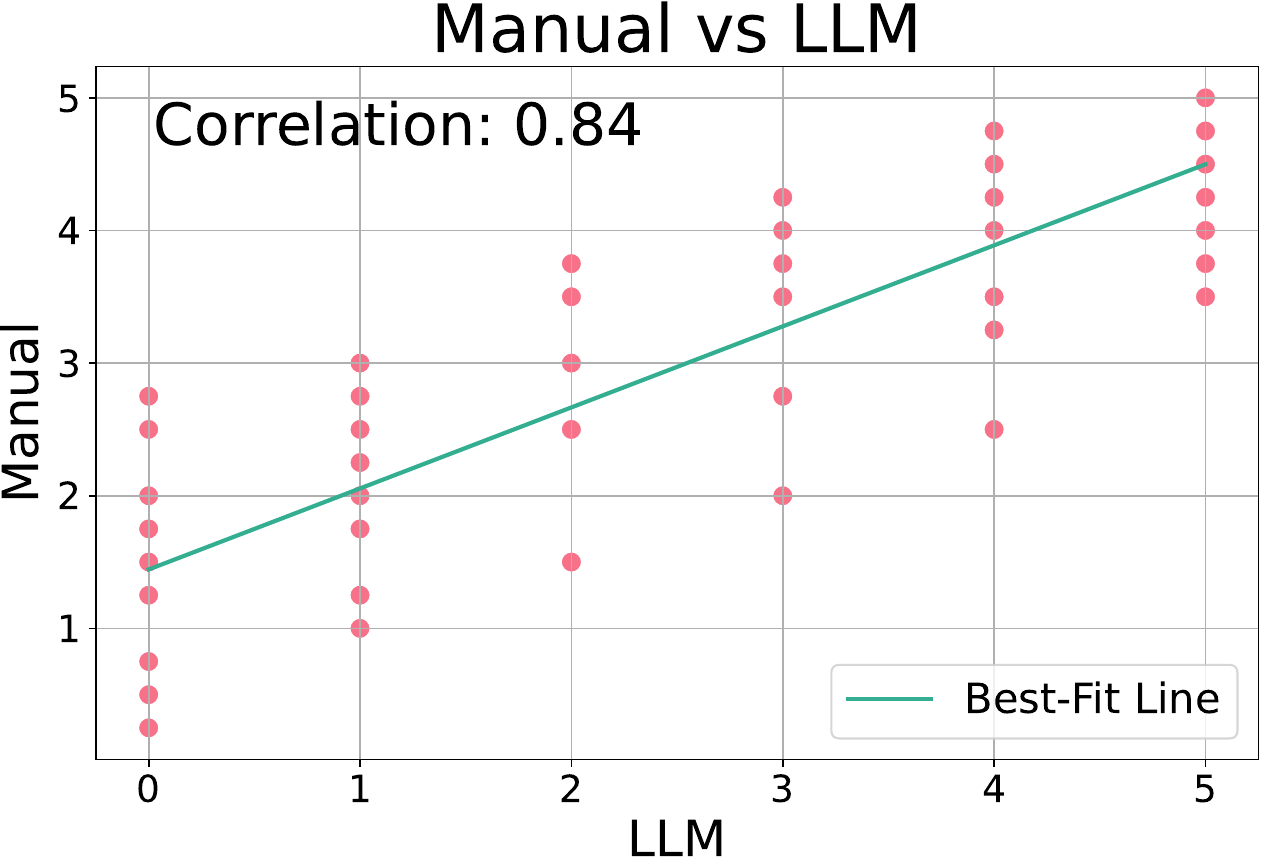}
    \hspace{0.1\textwidth}
    \includegraphics[width=0.32\linewidth]{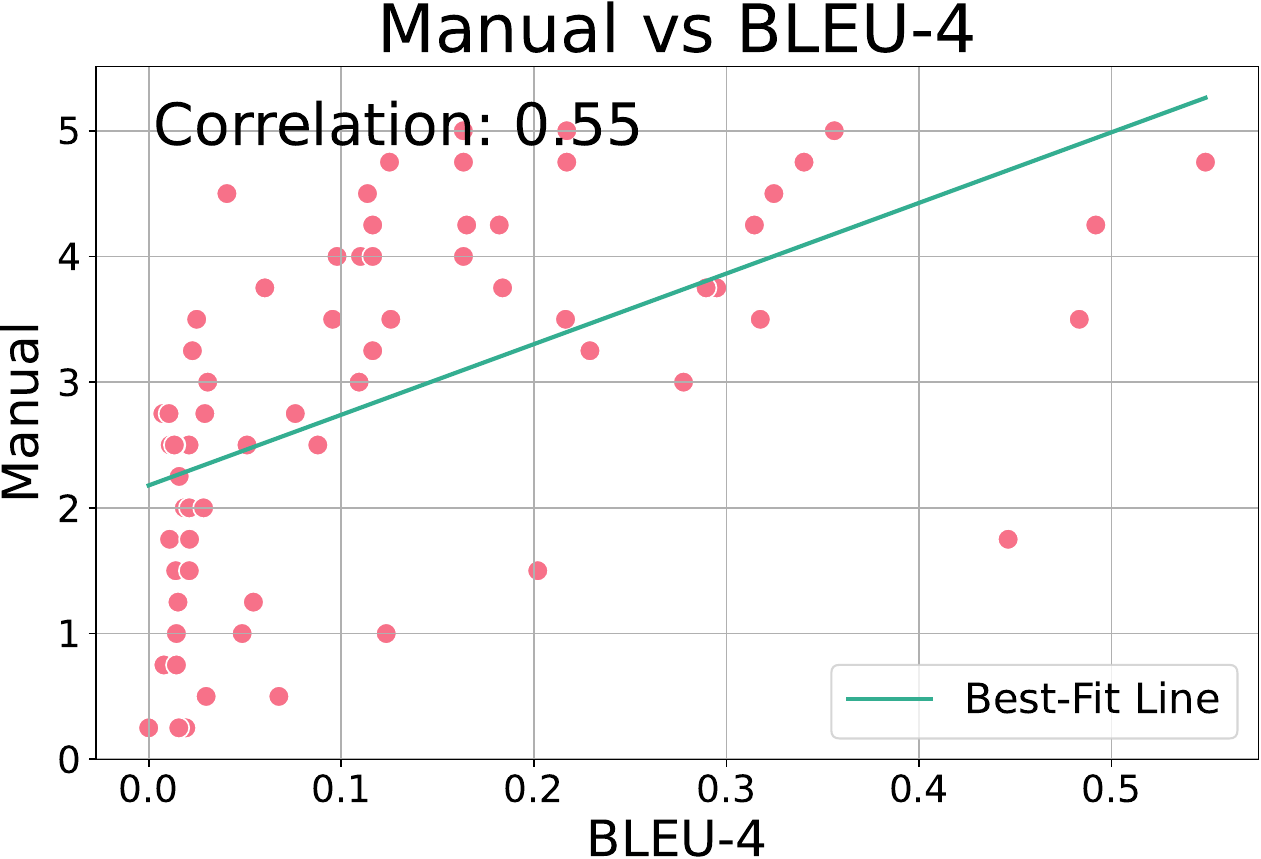} \\
    \vspace{0.3cm}
    \includegraphics[width=0.32\linewidth]{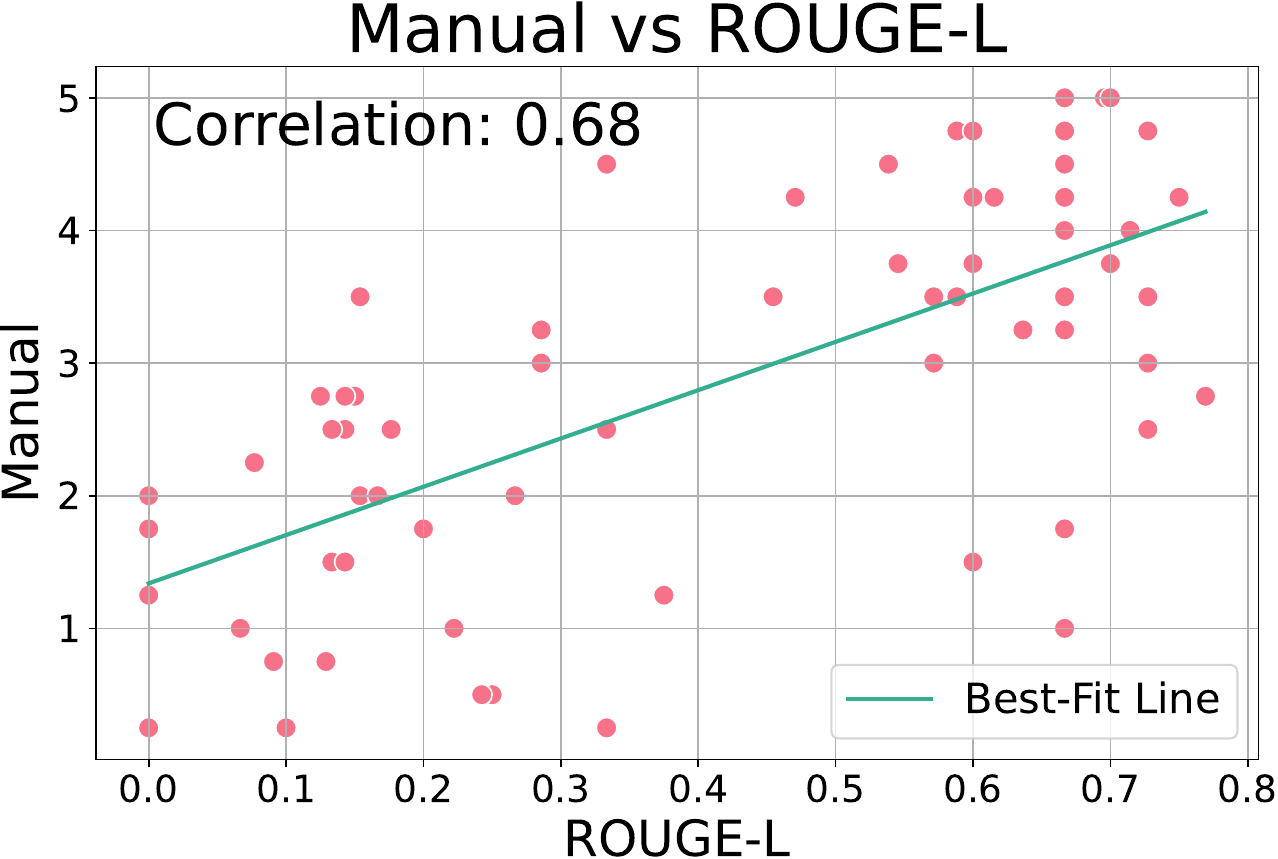}
    \hspace{0.1\textwidth}
    \includegraphics[width=0.32\linewidth]{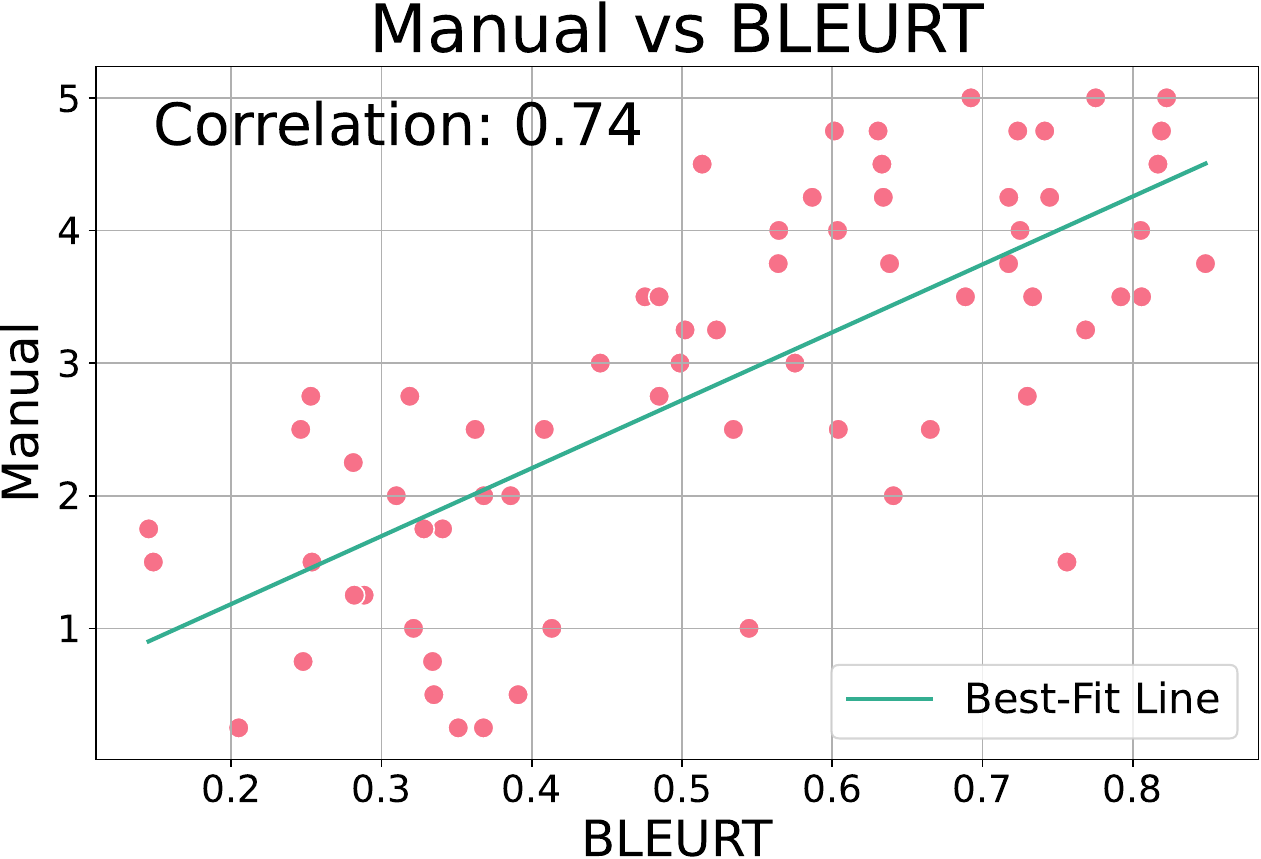}
    \vspace{-0.3cm}
    \caption{
        \textbf{Correlation between human judgement and evaluation metrics:} 
        We plot the scores obtained via human evaluation (`Manual') against the LLM evaluation scores and the standard captioning metrics (BLEU, ROUGE, and BLEURT). We observe that the LLM score correlates the most with human judgement. %
        }
    \label{fig:app:human_correlation}
\end{figure*}

\begin{figure*}
    \centering
    \includegraphics[width=0.32\linewidth]{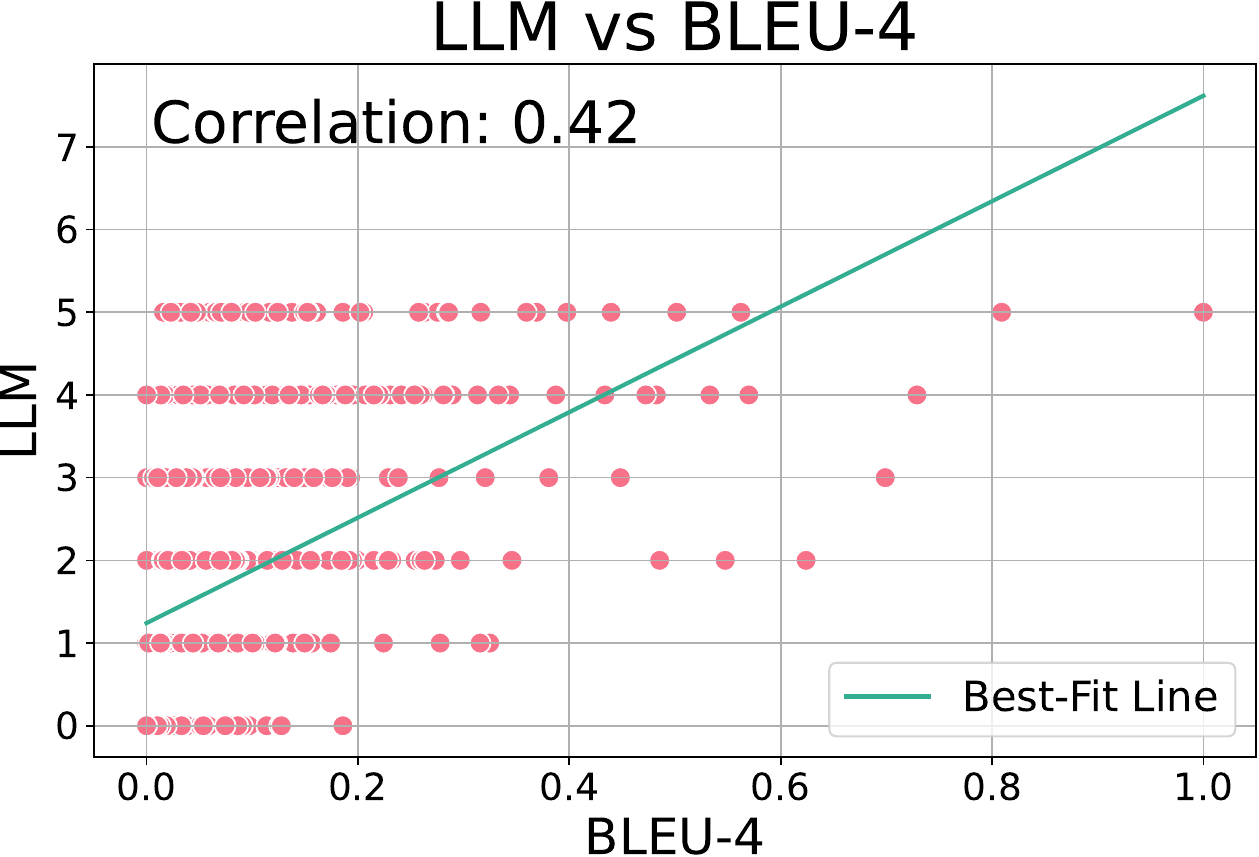}
    \includegraphics[width=0.32\linewidth]{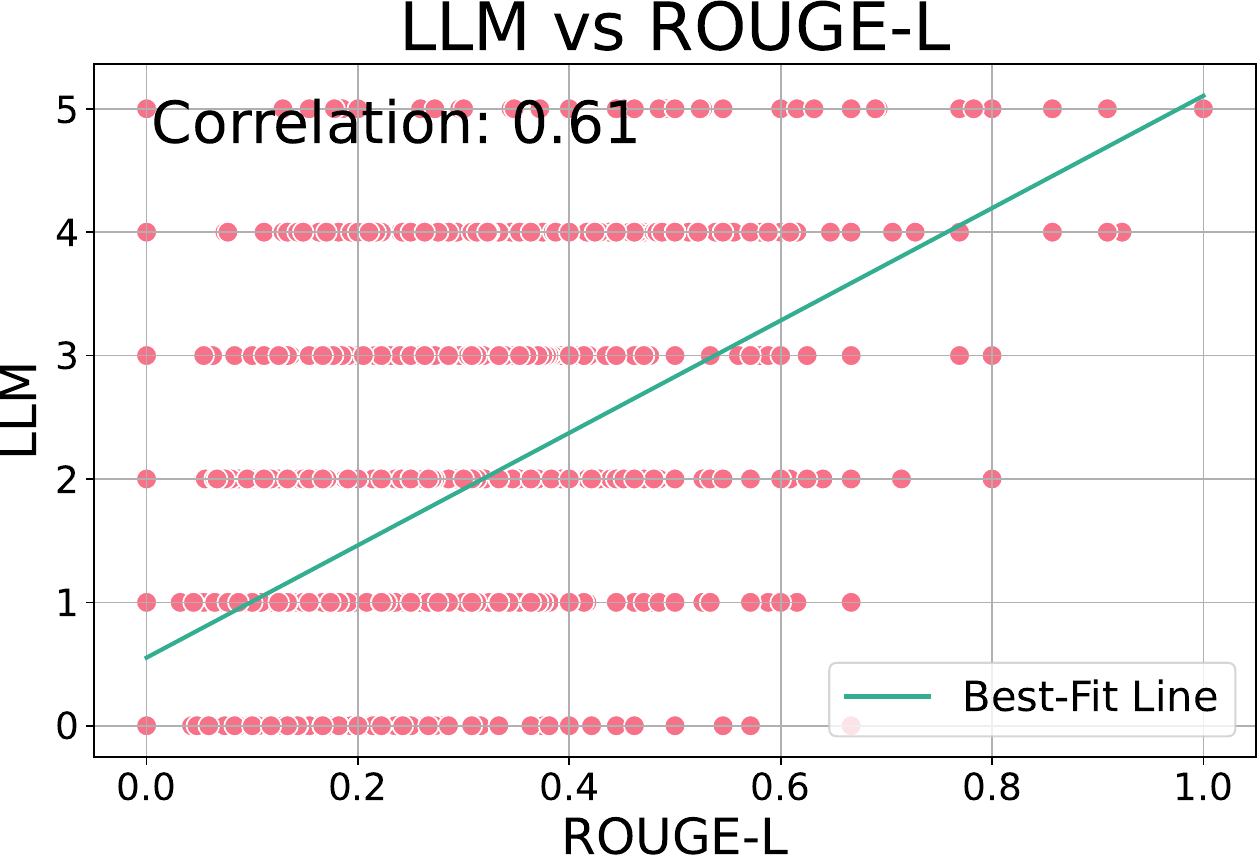}
    \includegraphics[width=0.32\linewidth]{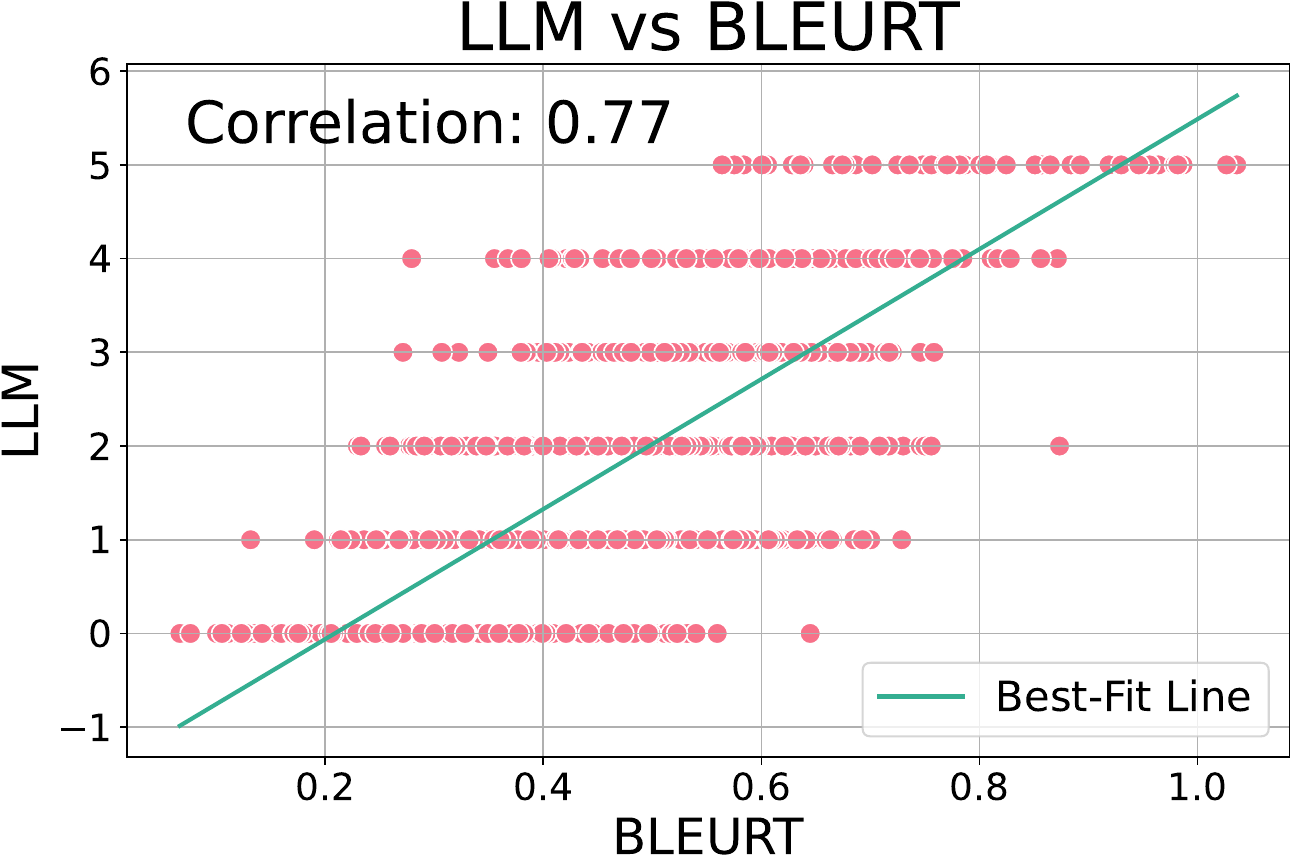}
    \vspace{-0.3cm}
    \caption{
        \textbf{Correlation between LLM evaluation and other metrics:} We
        plot the LLM scores on the BOBSL \valManual and compare against the standard captioning metrics.
    }
    \label{fig:app:LLM_eval_correlation}
\end{figure*}

\begin{figure}
    \centering
        \includegraphics[width=0.99\linewidth]{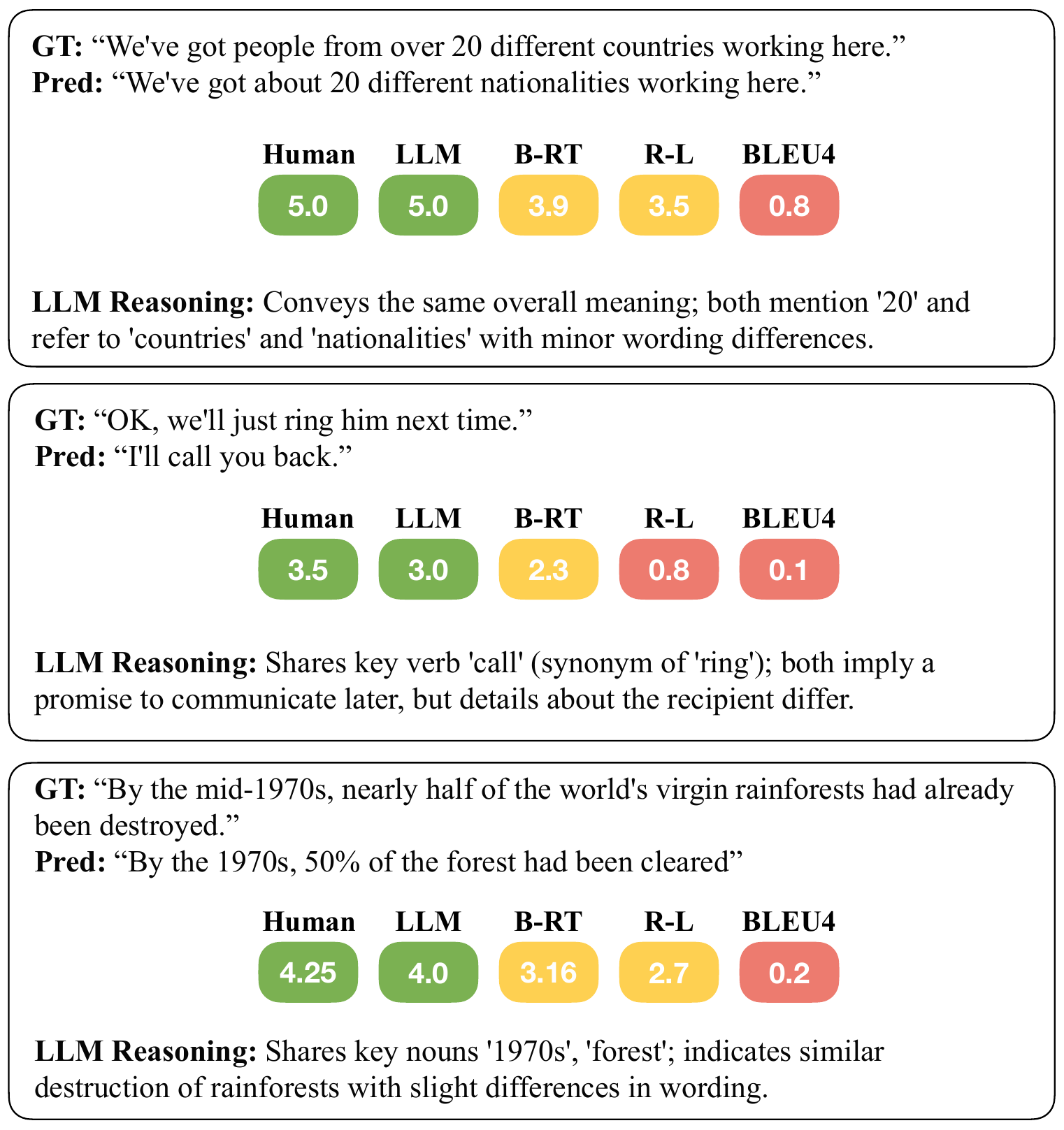}
    \vspace{-2mm}
    \caption{
        \textbf{Qualitative examples of LLM evaluation:} The scores obtained by LLM strongly correlate with human judgements while simultaneously being able to give detailed descriptions
        for the reasoning.
        All scores above are scaled to be out of 5 to make the comparison easier.
    }
    \label{fig:app:LLM_qualitative}
\end{figure}

\newpara{LLM evaluation analysis.}
As discussed in \appendixref{Sec.~4.3}{\cref{subsec:ablations}}, we provide additional analysis and statistics on LLM-based evaluation. A human study was conducted where 5 annotators manually scored a set of 70 translations. \cref{fig:app:human_correlation} and \cref{fig:app:LLM_eval_correlation} illustrate the correlation between the average of these human scores, various automatic metrics~\cite{bleu, sellam2020bleurt, lin-2004-rouge}, and our LLM-based evaluation metric.
As shown in \cref{fig:app:human_correlation}, the LLM-based metric exhibits the highest correlation with human judgments. 
This strong correlation highlights its potential as a useful method for evaluating sign language translations.
We further provide qualitative examples for the LLM scores in \cref{fig:app:LLM_qualitative}.

\input{figs/supmat/llmeval.tex}

\begin{table}
	\small
	\setlength{\tabcolsep}{9pt}
	\centering
	\resizebox{0.99\linewidth}{!}
	{
		\begin{tabular}{|m{2.5cm}|m{5cm}|} \hline
			\textbf{Type} & \textbf{Prompt} \\ \hline
			Initial & You are an AI assistant designed to interpret a video of a sign language signing sequence and translate it into English.
			\\ \hline
			Previous sentence & The previous context is the following: 
			\\ \hline
			Pseudo-glosses & The following are some possible words present in the sentence: 
			\\ \hline
			Background description & Description of the background is:
			\\ \hline
			Visual features & The following are the video tokens:
			\\ \hline
		\end{tabular}
	}
	\caption{
		\textbf{Prompt details.} Each cue is accompanied by a specific prompt that explains the task and helps the model differentiate between the various inputs.
	}
	\label{tab:app:prompt}
\end{table}

\subsection{Prompt details}
\label{subsec:app:prompts}
As explained in
\appendixref{Sec.~3.1}{\cref{subsec:overview}}
of the main paper,
we use five distinct prompts to define the
task and to describe each cue.
The exact 
prompts are provided in \cref{tab:app:prompt}.
Note that when randomly dropping a cue, we also omit the corresponding prompt.

\subsection{Background description collation}
\label{subsec:app:background}
\begin{figure*}
    \centering
    \includegraphics[width=1\linewidth]{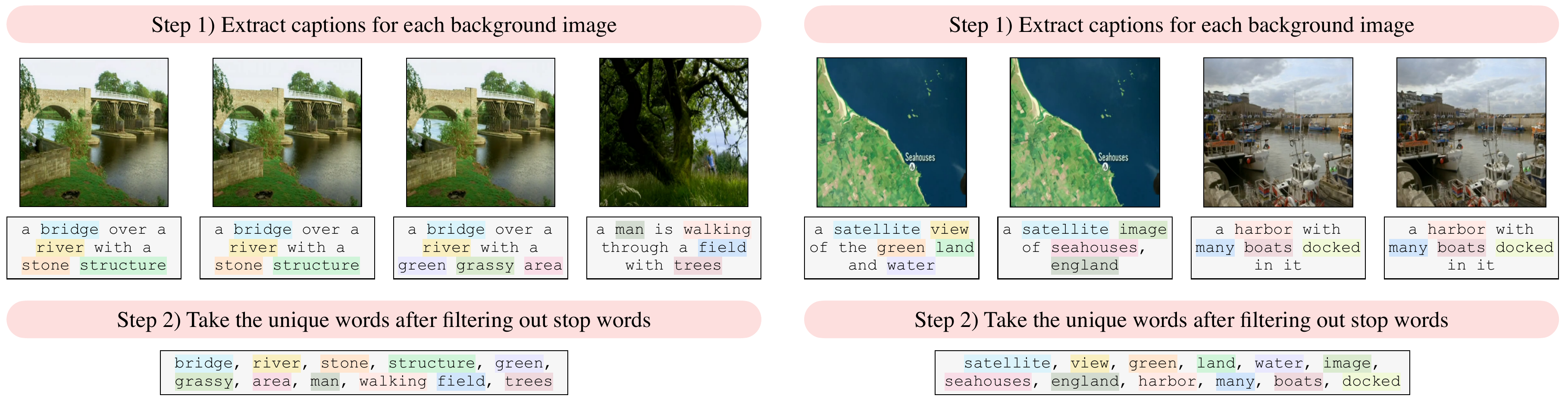}
	\caption{
		\textbf{Background description collection:} We illustrate, with two examples (left and right blocks), the process of collecting a background description from a signing sentence video. First, we use the BLIP2 image captioner~\cite{li2023blip2} to extract captions for a sequence of background frames in the video. Then, we remove stop words and use the set of unique words to represent the final background description.
	}
	\label{fig:app:bg}
\end{figure*}

In \cref{fig:app:bg}, we illustrate two examples
to show the process for the background description collation.
As explained in \appendixref{Sec.~3.2}{\cref{subsec:representation}}, in the first step,
we extract captions from multiple frames; in the second step, we take the unique words after filtering out stop words.

We further perform several analyses on these background descriptions on the BOBSL training set. First, we measure the similarity between background descriptions and the ground truth translation sentences, and obtain 3.4\% IoU, 5.3\% precision, and 9.3\% recall. We note that the informative signal in the background descriptions may be beyond the exact word overlap. Next, we look at the distribution of parts of speech, revealing 56.1\% nouns, 19.4\% verbs, 11.8\% adjectives, and 7.9\% proper nouns. Among the most frequently occurring words, ``man'' was identified as the most common noun, ``standing'' as the most common verb, and ``front'' as the most common adjective. %

\subsection{BOBSL mapping network}
\label{subsec:app:bobsl_mapping}
The details of the 2-layer MLP used as the mapping network are provided in \cref{tab:app:bobsl_mapping}. The input to the mapping network consists of Video-Swin features, and its output serves as the input to the LLM decoder. Specifically, in our experiments, the size of the Video-Swin features is 768, while the input size of the LLM decoder (Llama3-8B) is 4,096.

\begin{table}
	\scriptsize
	\centering
	\resizebox{0.55\columnwidth}{!}{
		\begin{tabular}{c|c|c}
			layer & input sizes & output sizes \\
			\hline
			{fc$_1$} & $T \times C$ & $T \times C'$ \\
			\hline
			{gelu}  & $T \times C'$ & $T \times C'$\\
			\hline
			{fc$_2$} & $T \times C'$ & $T \times C'$\\
	\end{tabular}
        }
	\caption{\textbf{Mapping network architecture for BOBSL training.}
		We display the 2-layer MLP details, which consists of fully-connected layers and a gelu activation. $C=768$ represents the number of channels in the Video-Swin features, while $C'=4,096$ denotes the input size of the LLM decoder. $T$ represents the temporal length of the input feature sequence, which has an average value of 56.
	}
	\label{tab:app:bobsl_mapping}
\end{table}

\begin{table} %
    \scriptsize
    \centering
    \resizebox{\columnwidth}{!}{
        \begin{tabular}{c|c|c|c|c|c}
            layer & kernel & stride & padding & input sizes & output sizes \\
            \hline
            {conv$_1$} & 5 & 1 & 2 & $T \times C$ & $T \times C$ \\
            \hline
            {relu$_1$} & - & - & - & $T \times C$ & $T \times C$ \\
            \hline
            {maxpool$_1$} & 2 & 2 & - & $T \times C$ & $T/2 \times C$ \\
            \hline
            {conv$_2$} & 5 & 1 & 2 & $T/2 \times C$ & $T/2 \times C$ \\
            \hline
            {relu$_2$} & - & - & - & $T/2 \times C$ & $T/2 \times C$ \\
            \hline
            {maxpool$_2$} & 2 & 2 & - & $T/2 \times C$ & $T/4 \times C$ \\
            \hline
            {fc$_1$} & - & - & - & $T/4 \times C$ & $T/4 \times C'$ \\
            \hline
            {gelu} & - & - & - & $T/4 \times C'$ & $T/4 \times C'$ \\
            \hline
            {fc$_2$} & - & - & - & $T/4 \times C'$ & $T/4 \times C'$ \\
        \end{tabular}}
    \caption{\textbf{Mapping network for How2Sign training.}
    	We apply 1D CNN on the visual features extracted from the Video-Swin ISLR model.
    	The output of this CNN is then fed into a 2-layer MLP. $C=768$ represents the number of channels in the Video-Swin features, while $C'=4,096$ denotes the input size of the LLM decoder. $T$ represents the temporal length of the input feature sequence, which has an average value of 171.
    }
    \label{tab:app:how2sign_mapping}
\end{table}

\subsection{How2Sign training details}
\label{subsec:app:H2S_training}
\newpara{ISLR training details.} 
As mentioned in \appendixref{Sec.~3.2}{\cref{subsec:representation}} of the main paper,
we fine-tune the Video-Swin model, which is released by~\cite{raude2024}, with annotations provided by~\cite{duarte22slretrieval}. The training data is automatically annotated from mouthing and dictionary sources, and we set thresholds at 0.75 and 0.5, respectively, to filter the data and enhance its reliability. 
We train the model on 4 A6000 GPUs with a batch size of 24 per GPU, utilising the Adam optimizer~\cite{KingmaAdam}. Training is performed in bfloat16 precision. The training spans 30 epochs, including the warmup phase for the first 1 epoch. The learning rate is set to 0.0001 and one cycle cosine learning rate scheduler is adapted.

\newpara{Visual features.}
We set the stride ($s$) to 1 for feature extraction using the Video-Swin model on the How2Sign dataset, as the data is smaller and manageable for training. The average number of features is 171, corresponding to a 6.8-second long sequence.

\newpara{Mapping network on How2Sign.} 
As mentioned in \appendixref{Sec.~3.3}{\cref{subsec:training}} of the main paper, we further provide detailed information about the mapping network for training our model on the How2Sign dataset.
Through our experiments, we found that training with only 2-layer MLP was not successful on the How2Sign dataset. Therefore, we add a simple 1D CNN before the MLP layers to compress the long sequences with minimal additional parameters. The 1D CNN is configured with a specific sequence of layers: \{K5, P2, K5, P2\}, where $K_{\sigma}$ denotes a kernel size of $\sigma$, and $P_{\sigma}$ represents a pooling layer with a kernel size of $\sigma$~\cite{hu2023continuous}.
Details of this mapping network, including input and output sizes, are provided in \cref{tab:app:how2sign_mapping}.

\begin{table}
    \small
    \setlength{\tabcolsep}{3pt}
    \centering
    \resizebox{0.75\linewidth}{!}
    {
    \begin{tabular}{l|c|cccccc}
        \toprule
        Model & Size  & B4 & B-RT & R-L & CIDEr & IoU & LLM \\
        \midrule
        Llama3.2 & 1B & 2.4 & 39.2 & 15.8 & 35.8 & 13.6 & 1.05 \\
        Llama3.2 & 3B & 3.1 & 40.0 & \textbf{17.0} & 41.9 & 14.6 & 1.20 \\
        \rowcolor{aliceblue} Llama3 & 8B & \textbf{3.3} & \textbf{40.3} & 16.9 & \textbf{41.9} & \textbf{14.8} & \textbf{1.20} \\
        \bottomrule
    \end{tabular}
    }
    \vspace{-2mm}
    \caption{
    \textbf{LLM decoder variants.}
    The Llama3-8B model used in the main paper performs overall better than
    more recent variants of Llama3.2 with less parameters. Note the results
    are reported on BOBSL \testManual.
    }
    \label{tab:app:llm}
\end{table}
    
\newpara{Target sentence augmentation for How2Sign.}
We observe overfitting starting from around 5–6 epochs when training with an LLM on the relatively small How2Sign dataset. To further improve the model's performance, we employ a data augmentation technique that randomly drops 0–20\% of the words from the GT sentences.

\section{Additional Experiments}
\label{sec:app:experiments}
We examine performance variations when using different LLM decoders (\cref{subsec:app:llm_variants}),
evaluate all possible cue combinations (\cref{subsec:app:combining_cues}),
investigate scenarios with missing cues (\cref{subsec:app:missing_cues}),
\todo{with multiple number of previous sentences (\cref{subsec:app:numprev}),
and with various background frame sampling rates (\cref{subsec:app:bgframes})}.
We also showcase the performance of our ISLR backbone on the HowSign dataset (\cref{subsec:app:islrhow2sign}),
and report the reproduction results of GFSLT~\cite{zhou2023gloss} and Sign2GPT~\cite{wong2024sign2gpt} on the PHOENIX14T dataset (\cref{subsec:app:phoenix}).
\todo{Finally, we demonstrate the applicability of our method on PHOENIX14T
(\cref{subsec:app:phoenixours}).}

\subsection{Llama decoder variants}
\label{subsec:app:llm_variants}
To further analyse the impact
of the LLM decoder on performance, we
experiment with
various Llama variants.
Specifically, we compare the Llama3-8B model used in the main paper experiments to more recent
and smaller Llama3.2 models:
Llama3.2-1B and Llama3.2-3B.
As shown in \cref{tab:app:llm}, Llama3.2-3B demonstrates performance comparable to Llama3-8B. When using the Llama3.2-1B model, we observe a performance drop of 0.7 in the BLEU-4 (B4) score compared to the Llama3-8B model. However, Llama3.2-1B still outperforms all baselines compared in \appendixref{Tab.~3}{\cref{tab:sota-bobsl}} of the main paper. Note that this experiment is conducted on BOBSL \testManual.

\begin{table}
    \small
    \setlength{\tabcolsep}{9pt}
    \centering
    \resizebox{0.9\linewidth}{!}
    {
        \begin{tabular}{cccc|lll}
        \toprule
        Vid & PG & \PredPrev  & BG & B-RT & IoU & LLM  \\
        \midrule
        \checkmark & & & & 41.0 & 16.6 & 1.29\\
        \midrule
        \checkmark & \checkmark & & & 41.8 & 17.5 & 1.40\\
        \checkmark & & \checkmark & & 41.5 & 17.0 & 1.38 \\
        \checkmark & & & \checkmark & 41.9 & 17.5 & 1.41 \\
        \midrule
        \checkmark & \checkmark & \checkmark  & & 42.5 & 18.1 & 1.45\\

        \checkmark & \checkmark & & \checkmark & 43.2 & 18.6 & 1.54 \\
        \checkmark & & \checkmark & \checkmark & 43.1 & 18.3 & 1.52 \\
        \midrule
        \rowcolor{aliceblue} \checkmark & \checkmark & \checkmark  &  \checkmark & \textbf{43.5} & \textbf{18.8} & \textbf{1.56} \\
        \bottomrule
        \end{tabular}
    }
    \vspace{-2mm}
    \caption{
    \textbf{Combining different cues.}
    We complement \appendixref{Tab.~1}{\cref{tab:combining-cues}} of the main paper with more combination of cues and report results on BOBSL \valManual. A checkmark $\checkmark$ indicates a cue provided during training and testing. %
    }
    \label{tab:app:combining-cues-sup}
\end{table}

\subsection{Combining different cues}
\label{subsec:app:combining_cues}
We complement \appendixref{Tab.~1}{\cref{tab:combining-cues}} of the main paper by providing results of all possible cue combinations in \cref{tab:app:combining-cues-sup}. 
These experiments reveal consistent performance improvements with each added cue, demonstrating that all cues complement each other.

\subsection{Missing cue scenario}
\label{subsec:app:missing_cues}
As discussed in \appendixref{Sec.~3.3}{\cref{subsec:training}} of the main paper, the \textit{Drop Cue} augmentation enables our model to perform flexible translations even when certain cues are missing during test time. The experimental results are presented in \cref{tab:app:missing-cue}. Note that, while the previous \cref{tab:app:combining-cues-sup} displays the performance of models trained with various cue combinations, \cref{tab:app:missing-cue} reports the inference results of the model trained with \textit{all} cues. Notably, our final model achieves results comparable to those of the models trained on specific combinations of cues (i.e.\ models listed in \appendixref{Tab.~1}{\cref{tab:combining-cues}} of the main paper).
This demonstrates that our final model can perform sign language translation with minimal performance degradation when certain cues are unavailable during inference.

\begin{table}
    \small
    \setlength{\tabcolsep}{9pt}
    \centering
    \resizebox{0.9\linewidth}{!}
    {
        \begin{tabular}{cccc|lll}
        \toprule
        Vid & PG & \PredPrev  & BG & B-RT & IoU & LLM  \\
        \midrule
        \checkmark & & & & 41.1 & 17.2 & 1.31 \\
        \midrule
        \checkmark & \checkmark &   & & 41.8 & 18.1 & 1.41 \\
        \checkmark & & \checkmark & & 41.7 & 17.3 & 1.39 \\
        \checkmark & & & \checkmark & 42.0 & 17.0 & 1.42 \\
        \midrule
        \checkmark & \checkmark & \checkmark & & 42.6 & 17.8 & 1.49 \\
        \checkmark & \checkmark & & \checkmark & 42.8 & 18.7 & 1.52 \\
        \checkmark & & \checkmark & \checkmark & 42.5 & 17.7 & 1.52 \\
        \midrule
        \rowcolor{aliceblue} \checkmark & \checkmark & \checkmark & \checkmark & 43.5 & 18.8 & 1.56 \\

        \bottomrule
        \end{tabular}
    }
    \caption{
    \textbf{Missing cue scenario at test time.} We perform inference using the model trained with all cues. A checkmark $\checkmark$ indicates a cue provided during inference, while a blank space denotes a missing cue. Results are reported on BOBSL \valManual. %
    }
    \label{tab:app:missing-cue}
\end{table}

\begin{table}
	\centering
	\resizebox{0.85\linewidth}{!}
	{
		\begin{tabular}{l c c c c c c}
			\toprule
			\#prev. & B4 & B-RT & R-L & CIDEr & IoU & LLM \\
			\midrule
		     1 & 3.3 & 40.3 & 16.9 & 41.9 & 14.8 & 1.20 \\
		 2 & 3.3 & 40.6 & 17.0 & 42.8 & 14.9 & 1.21 \\
			3 & 3.4 & 41.0 & 17.2 & 43.9 & 15.1 & 1.24 \\
			\bottomrule
		\end{tabular}
	}
    \vspace{-0.3cm}
	\caption{\todo{\textbf{Number of previous sentences (BOBSL \testManual).} We experiment with giving more previous context, and achieve only marginal improvements.}
	}
	\label{tab:app:previous}
\end{table}

\subsection{Number of previous sentences}
\label{subsec:app:numprev}
\todo{\cref{tab:app:previous} reports the results of our best model (Vid+PG+\PredPrev+BG) on \testManual, using 2–3 previous sentences as context during both training and inference. As shown in the table, providing a longer previous context results in only a marginal improvement (+0.1 BLEU-4 and +0.3 ROUGE). While additional context slightly improves B-RT (41.0 vs. 40.3), it comes at the cost of increased computational overhead during inference. To balance performance and efficiency, we use only a single previous sentence for inference.}

\begin{table}
	\centering
	\resizebox{0.99\linewidth}{!}
	{
		\begin{tabular}{c c c c c c c}
			\toprule
			Sampling rate (sec) & B4 & B-RT & R-L & CIDEr & IoU & LLM \\
			\midrule
		     1 & 3.3 & 40.3 & 16.9 & 41.9 & 14.8 & 1.20 \\
		 2 & 3.3 & 40.2 & 17.0 & 41.3 & 14.8 & 1.20 \\
			3 & 3.1 & 40.0 & 16.9 & 40.9 & 14.5 & 1.19 \\
			\bottomrule
		\end{tabular}
	}
    \vspace{-0.3cm}
    \caption{\todo{
    \textbf{Background frame sampling rate (BOBSL \testManual).} We experiment with the background frame sampling rate, by sampling a caption every 1-2-3 seconds, and see little effect on performance when sampling less frames (last row).
    Note that each sentence lasts on average 4.5 seconds.} %
	}
	\label{tab:app:sample}
\end{table}

\subsection{Background frame sampling rate}
\label{subsec:app:bgframes}

\todo{\cref{tab:app:sample} reports the results of our best model (Vid+PG+\PredPrev+BG) on \testManual, when varying the sampling rate of the background frames
during both training and inference. 
In the rest of the experiments, we sample a background caption every 1~second.
In \cref{tab:app:sample}, we experiment with reducing this rate by sampling every 2 or 3 seconds.
Since we remove repeated words in background captions,
we do not experiment with sampling more than 1~frame per second
(adjacent frames often depict nearly identical scenes). %
We obtain (40.3, 40.2, 40.0) BLEURT when sampling every (1, 2, 3) seconds, indicating little effect on performance at lower sampling rates. Thus, sampling every second avoids missing scene transitions while remaining computationally feasible.}

\begin{table}
    \small
    \setlength{\tabcolsep}{3pt}
    \centering
    \resizebox{0.99\linewidth}{!}
    {
        \begin{tabular}{l|l|cc|cc}
        \toprule
        Model & Training & \multicolumn{2}{c|}{Per-instance} & \multicolumn{2}{c}{Per-class} \\
        & & top-1 & top-5 & top-1 & top-5 \\
        \midrule
        I3D~\cite{duarte22slretrieval} & BOBSL $\rightarrow$ How2Sign & 59.5 & 78.9 & 44.5 & 68.7 \\ 
        Video-Swin (Ours) & How2Sign & 63.9 & 86.0 & 41.8 & 69.3 \\ 
        Video-Swin (Ours) & BOBSL $\rightarrow$ How2Sign & \textbf{77.0} & \textbf{92.8} & \textbf{58.5} & \textbf{82.3} \\ 
        \bottomrule
        \end{tabular}
    }
    \vspace{-0.3cm}
    \caption{
    \textbf{ISLR performance on How2Sign test set.} Per-instance accuracy is measured over all test instances, while per-class accuracy reflects the average performance across the sign categories in the test set. 
    }
    \label{tab:app:islr_performance}
\end{table}

\subsection{ISLR performance on How2Sign}
\label{subsec:app:islrhow2sign}
The test set provided by~\cite{duarte22slretrieval} is composed of 2,212 manually annotated data. We evaluate both per-instance and per-class accuracy metrics. Per-instance accuracy is calculated across all test instances, while per-class accuracy represents the average performance across the sign categories in the test set. This metric is particularly useful for addressing the unbalanced nature of the datasets, as recommended in~\cite{Albanie20}.

As shown in \cref{tab:app:islr_performance}, our Video-Swin ISLR model, trained without pre-training on the BOBSL dataset, achieves performance comparable to the I3D ISLR model~\cite{duarte22slretrieval}, which is pre-trained on the BOBSL dataset and fine-tuned on the How2Sign dataset.
Furthermore, when the Video-Swin ISLR model is initialised with weights pre-trained on the BOBSL dataset, as released by~\cite{raude2024}, and further fine-tuned on the How2Sign dataset by us, it achieves a 13.1\% improvement in per-instance top-1 accuracy and a 16.7\% improvement in per-class top-1 accuracy. This underscores the effectiveness and robustness of our framework's ISLR backbone.

\subsection{Reproducing GFSLT and Sign2GPT on PHOENIX14T}
\label{subsec:app:phoenix}
As mentioned in
\appendixref{Sec.~4.2}{\cref{subsec:baselines}}
of the main paper, we reproduce the performance of the GFSLT and Sign2GPT models on the PHOENIX14T dataset \cite{Phoenix,camgoz-slt}. The results are shown in \cref{tab:app:reproduce}. The $\dagger$ symbol denotes the reproduced results, which show comparable performance to the results reported in their original papers across all metrics.

\begin{table}
    \small
    \setlength{\tabcolsep}{3pt}
    \centering
    \resizebox{0.9\linewidth}{!}
    {
        \begin{tabular}{l|ccccc}
        \toprule
        Model  & B1 & B2 & B3 & B4 & R-L \\
        \midrule
        GFSLT~\cite{zhou2023gloss} & 43.71 & 33.18 & 26.11 & 21.44 & 42.49 \\ 
        GFSLT~\cite{zhou2023gloss} $\dagger$ & 42.02 & 31.88 & 25.30 & 20.76 & 42.62 \\ 
        \midrule
        Sign2GPT~\cite{wong2024sign2gpt} & 45.43 & 32.03 & 24.23 & 19.42 & 45.23 \\ 
        Sign2GPT~\cite{wong2024sign2gpt} $\dagger$ & 44.14 & 32.72 & 25.49 & 20.82 & 43.70 \\ 
        \midrule
        Sign2GPT (w/PGP)~\cite{wong2024sign2gpt} & 49.54 & 35.96 & 28.83 & 22.52 & 48.90 \\ 
        Sign2GPT (w/PGP)~\cite{wong2024sign2gpt} $\dagger$ & 46.90 & 35.72 & 28.30 & 23.22 & 46.28 \\ 

        \bottomrule
        \end{tabular}
    }
    \vspace{-0.3cm}
    \caption{
    \textbf{Reproducing GFSLT and Sign2GPT on PHOENIX14T.} $\dagger$ denotes our reproduction results and PGP denotes pseudo-gloss pre-training introduced in~\cite{wong2024sign2gpt}.
    }
    \label{tab:app:reproduce}
\end{table}

\begin{table}%
    \small
    \setlength{\tabcolsep}{3pt}
    \centering
    \resizebox{0.9\linewidth}{!}
    {
        \begin{tabular}{l|ccccc}
        \toprule
        Model  & B4 & B-RT & R-L & CIDEr & IoU \\
        \midrule
        GFSLT~\cite{zhou2023gloss} & 21.44 & - & 42.49 & - & - \\ Sign2GPT~\cite{wong2024sign2gpt} & 19.42 & - & 45.23 & - & - \\
        Sign2GPT (w/PGP)~\cite{wong2024sign2gpt} & 22.52 & - & \textbf{48.90} & - & - \\
        Ours (Vid) & 20.58 & 52.25 & 41.20 & 190.40 & 34.05 \\ 
        Ours (Vid+PG~\cite{ahn2024slowfast}) & \textbf{23.80} & \textbf{52.80} & 46.11 & \textbf{227.05} & \textbf{38.49} \\ 
        \bottomrule
        \end{tabular}
    }
    \vspace{-0.3cm}
    \caption{
    \textbf{Evaluation of our model on PHOENIX14T test split.} \todo{Incorporating PG as an additional textual cue improves performance across all evaluation metrics.}
    }
    \label{tab:app:phoenix14t}
\end{table}

\newpara{Training on BOBSL.}
For GFSLT, we observed that using the official codebase leads to gradient divergence during the masked word reconstruction process in text decoding. To mitigate this issue, we reduced the weight of the word reconstruction loss from 1 to 0.1. For Sign2GPT, as the official codebase only includes the model and hyperparameters, we developed training code using \texttt{Accelerate}~\cite{accelerate} framework.

\begin{table*}
	\setlength{\tabcolsep}{9pt}
	\resizebox{1\linewidth}{!}
	{
		\begin{tabular}{lll}
			\toprule
			\multirow{4}{*}{\textbf{1}} & Reference: & It's blind to the genius loci. \\
			& Candidate: & And that's what it means to be dislocated. \\
			& Score: & 0 \\
			& Reason: & No shared key nouns or verbs; the reference mentions `blind' and `genius loci', while the candidate mentions `dislocated'; meanings are different. \\
			\midrule
			\multirow{4}{*}{\textbf{2}} & Reference: & She put it by the entrance to the earth so we figure that they like heavy metal or something. \\
			& Candidate: & You've been in a wheelchair for a long time. \\
			& Score: & 0 \\
			& Reason: & No shared key nouns or verbs; the reference talks about `entrance', `earth', `heavy metal', while the candidate mentions `wheelchair'; meanings are unrelated. \\
			\midrule
			\multirow{4}{*}{\textbf{3}} & Reference: & You're coming along to the finale tomorrow? \\
			& Candidate: & I'll have to wait until tomorrow. \\
			& Score: & 1 \\
			& Reason: & Shares the key noun `tomorrow' but lacks other key content; meanings are somewhat related but differ. \\
			\midrule
			\multirow{4}{*}{\textbf{4}} & Reference: & A man's can was open for attack at any point in their life. \\
			& Candidate: & It's not a joke, it's a way of life. \\
			& Score: & 1  \\
			& Reason: & Shares the key noun `life', but overall meanings are different; reference discusses vulnerability, candidate discusses lifestyle. \\
			\midrule
			\multirow{4}{*}{\textbf{5}} & Reference: & Richard called English Nature, who told him that they were natterjack toads, Britain's rarest amphibian. \\
			& Candidate: & Richard, a Nottinghamshire Englishman, is a naturalist. \\
			& Score: & 2 \\
			& Reason: & Shares key nouns `Richard' and `English'; candidate omits details about `natterjack toads' and `Britain's rarest amphibian. \\
			\midrule
			\multirow{4}{*}{\textbf{6}} & Reference: & Cromwell treated Ireland like the primitive colony he thought it was, moving the Irish off their farms and using the land to pay his soldiers. \\
			& Candidate: & Cromwell was just one of many areas where the IRA set up roadblocks to stop loyalist paramilitaries and farmers from getting through. \\
			& Score: & 2 \\
			& Reason: & Shares key nouns `Cromwell', `Ireland', `farmers'; reference discusses historical actions, candidate discusses modern events; meanings differ. \\
			\midrule
			\multirow{4}{*}{\textbf{7}} & Reference: & He sort of guessed it would be 21 maybe 28 days, ended-up being 35. \\
			& Candidate: & He thought it was 21 days. \\
			& Score: & 3 \\
			& Reason: & Matches key nouns `he', `thought', `21 days'; candidate conveys a similar time estimation with fewer details. \\
			\midrule
			\multirow{4}{*}{\textbf{8}} & Reference: & They get as high as they can off the ground to push the other male down to the floor, and once that male becomes subservient, he slinks off and the dominant \\
			& & male then goes to his female to breed. \\
			& Candidate: & But when one of them is selected, it's the females that can be changed to breed with the new male. \\
			& Score: & 3 \\
			& Reason: & Shares key nouns `male', `female', `breed'; both discuss breeding behaviors, though specifics differ. \\
			\midrule
			\multirow{4}{*}{\textbf{9}} & Reference: & It's a lack of understanding on both sides. \\
			& Candidate: & I don't understand that. \\
			& Score: & 3 \\
			& Reason: & Shares key verb `understand'; both express lack of understanding; candidate is less specific. \\
			\midrule
			\multirow{4}{*}{\textbf{10}} & Reference: & OK, we'll just ring him next time. \\
			& Candidate: & I'll call you back. \\
			& Score: & 4 \\
			& Reason: & Shares key verb `call' (synonym of `ring'); both involve making a call; minor differences in context. \\
			\midrule
			\multirow{4}{*}{\textbf{11}} & Reference: & Really excited. \\
			& Candidate: & I'm so excited. \\
			& Score: & 5 \\
			& Reason: & Conveys the same overall meaning; both express excitement with minor wording differences.\\
			\midrule
			\multirow{4}{*}{\textbf{12}} & Reference: & Every day is totally different. \\
			& Candidate: & You know, every day is different. \\
			& Score: & 5 \\
			& Reason: & Conveys the same overall meaning; both state that each day is different with minor phrasing differences. \\
			\bottomrule
		\end{tabular}
	}
	\vspace{-0.3cm}
	\caption{\textbf{LLM evaluation in-context examples:} We display the set of 12 in-context examples provided to \texttt{GPT-4o-mini} \cite{gpt4} to calibrate the LLM evaluation metric. Each reference-candidate pair is provided to the LLM in the user prompt, with the expected output (score and reason) being provided with the assistant role as shown in \cref{fig:app:llmevalprompt}.
	}
        \vspace{-0.5cm}
	\label{tab:app:incontext}
\end{table*}

\subsection{Evaluation of our model on PHOENIX14T}
\label{subsec:app:phoenixours}
\todo{To evaluate the generalisability of our model, we conduct experiments on the PHOENIX14T dataset~\cite{Phoenix,camgoz-slt}. We extract video features using a Video-Swin model trained with pseudo-glosses (PG) obtained from SlowFastSign~\cite{ahn2024slowfast}.
As shown in \cref{tab:app:phoenix14t}, when fine-tuning an LLM-based model using only video features, we achieve a BLEU-4 score of 20.58 and a ROUGE-L score of 41.20. By incorporating PG as an additional textual cue, performance improves to a BLEU-4 score of 23.80 and a ROUGE-L score of 46.11. Compared to Sign2GPT~\cite{wong2024sign2gpt}, which reports a BLEU-4 score of 22.52 and a ROUGE-L score of 48.90, our model demonstrates comparable performance while highlighting the effectiveness of leveraging PG as an additional cue.

Note that no other contextual information (i.e.\ previous sentence, background) is available for this evaluation.
Although PHOENIX14T consists of TV weather broadcasts, the dataset is segmented at the sentence level and lacks context from previous sentences or background information. Moreover, the dataset includes manually annotated glosses and has a restricted vocabulary, leading to performance saturation.
This setting therefore does not reflect open-vocabulary tasks, which present greater challenges in real-world scenarios.}

\section{Additional Qualitative Results}
\label{sec:app:qualitative}

\begin{figure*}
    \centering
        \includegraphics[width=1\linewidth]{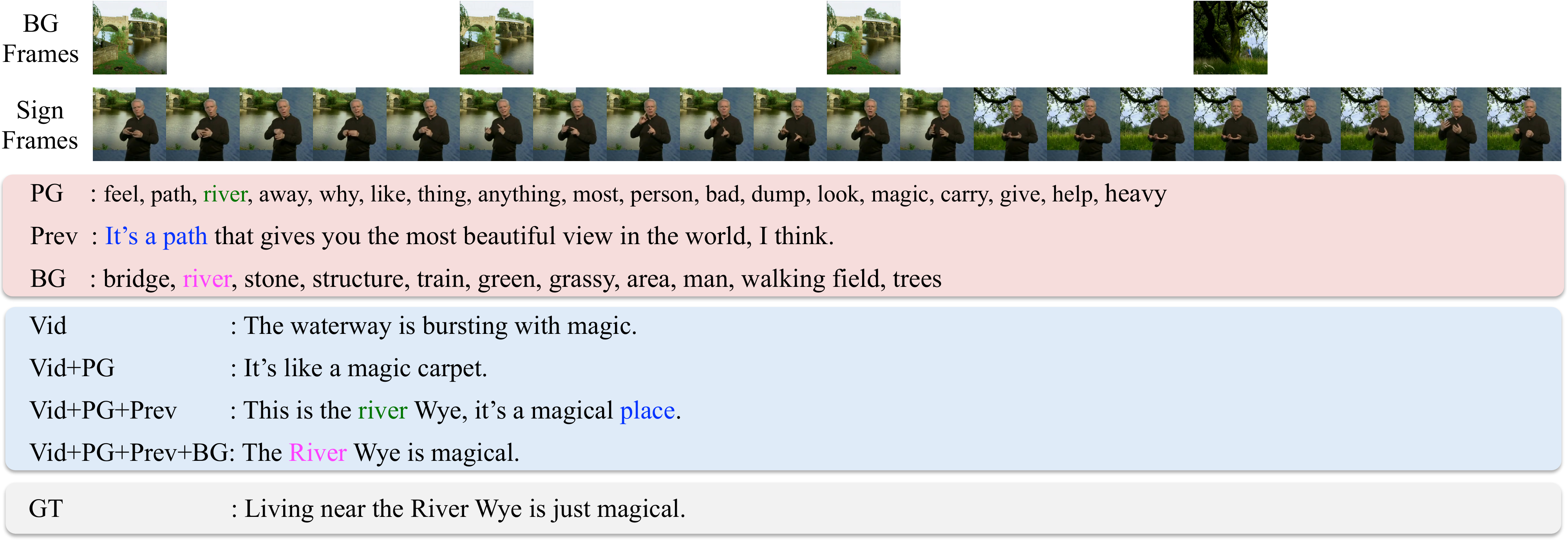}\vspace{0.1cm}
        \includegraphics[width=1\linewidth]{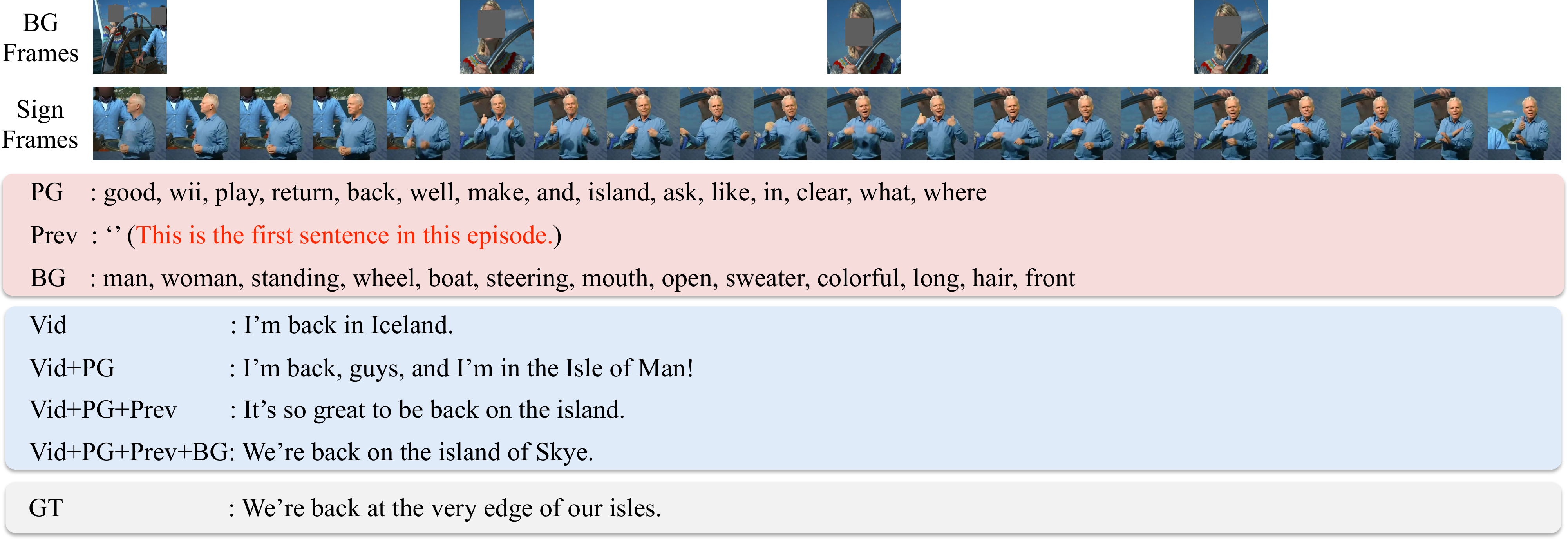}\vspace{0.1cm}
        \includegraphics[width=1\linewidth]{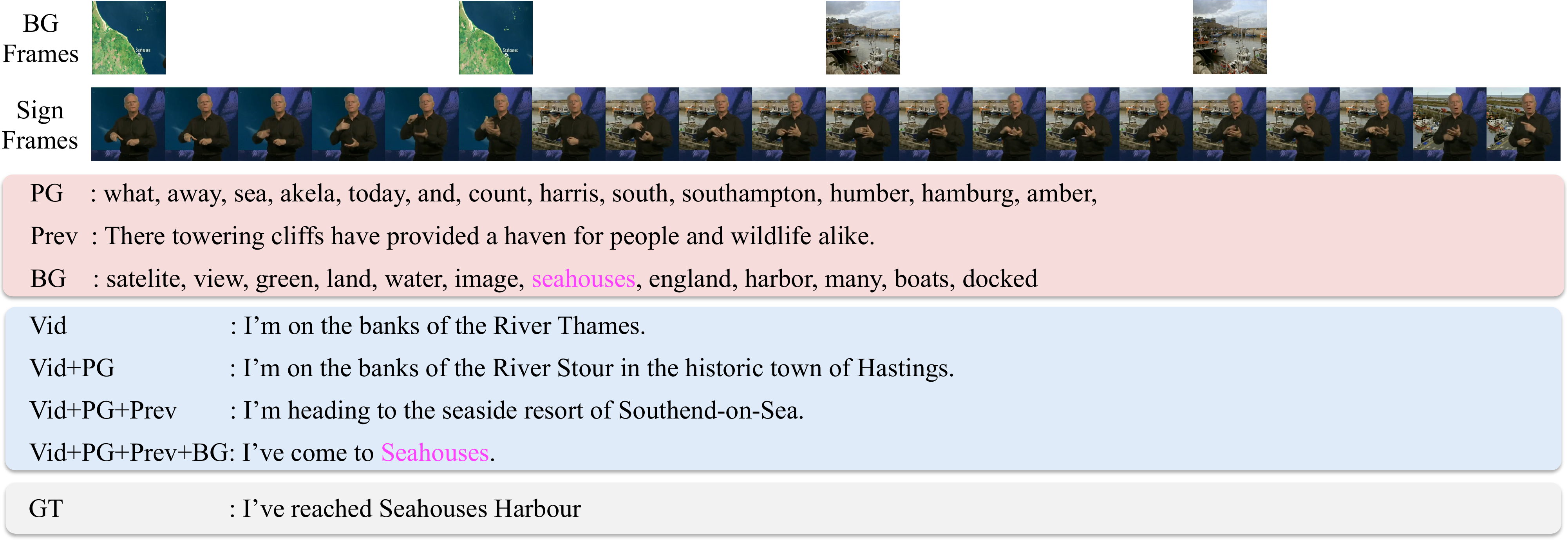}
    \caption{
        \textbf{Qualitative results:}
        We complement \appendixref{Fig.~3}{\cref{fig:qualitative}} of the main paper with more examples.
    }
    \label{fig:app:qualitative1}
\end{figure*}

\begin{figure*}
    \centering
    \includegraphics[width=1\linewidth]{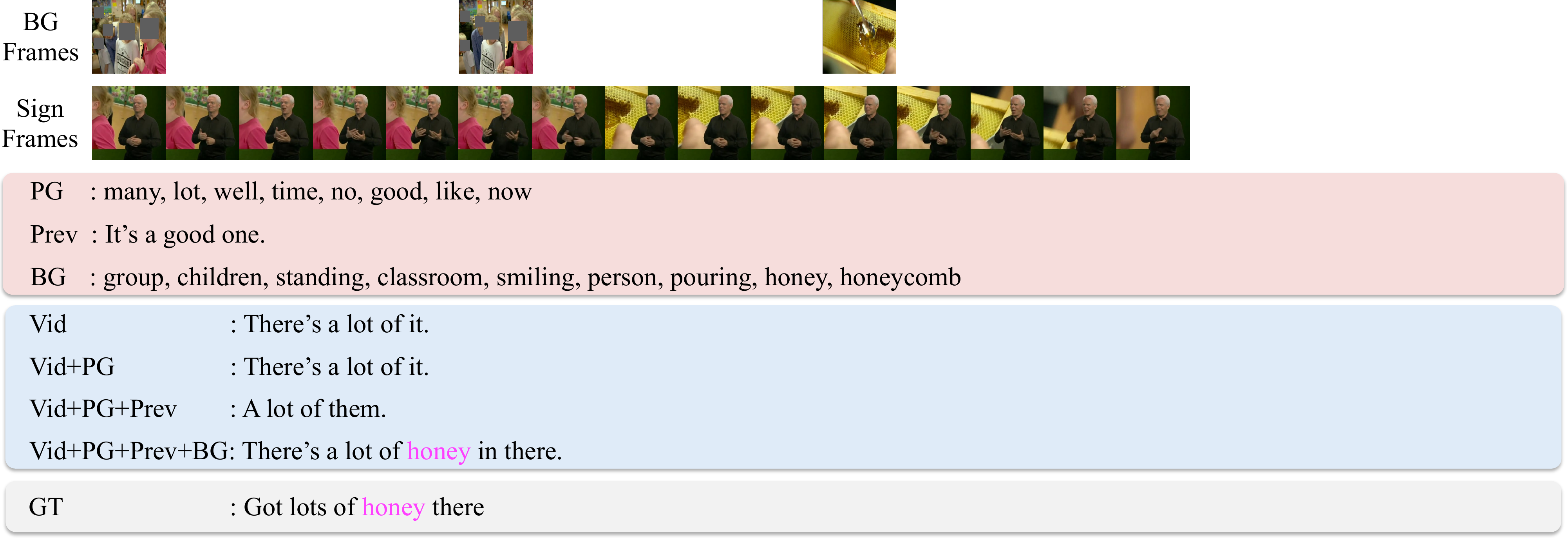}\vspace{0.1cm}
    \includegraphics[width=1\linewidth]{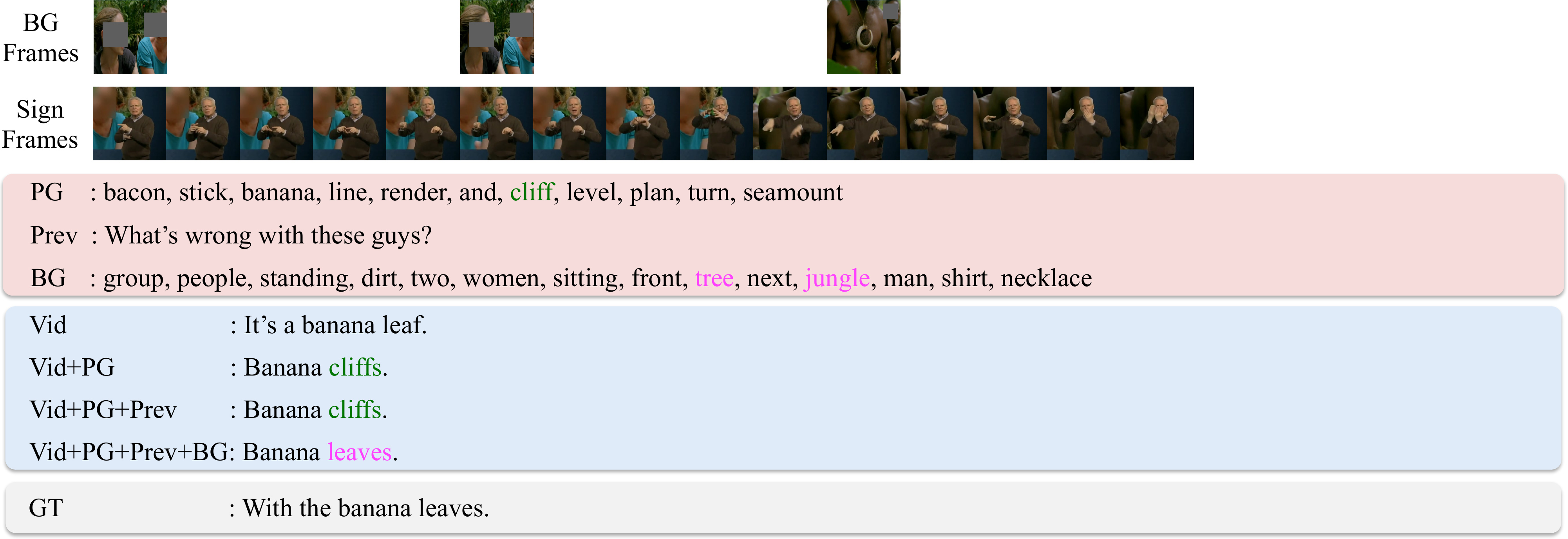}\vspace{0.1cm}
    \includegraphics[width=1\linewidth]{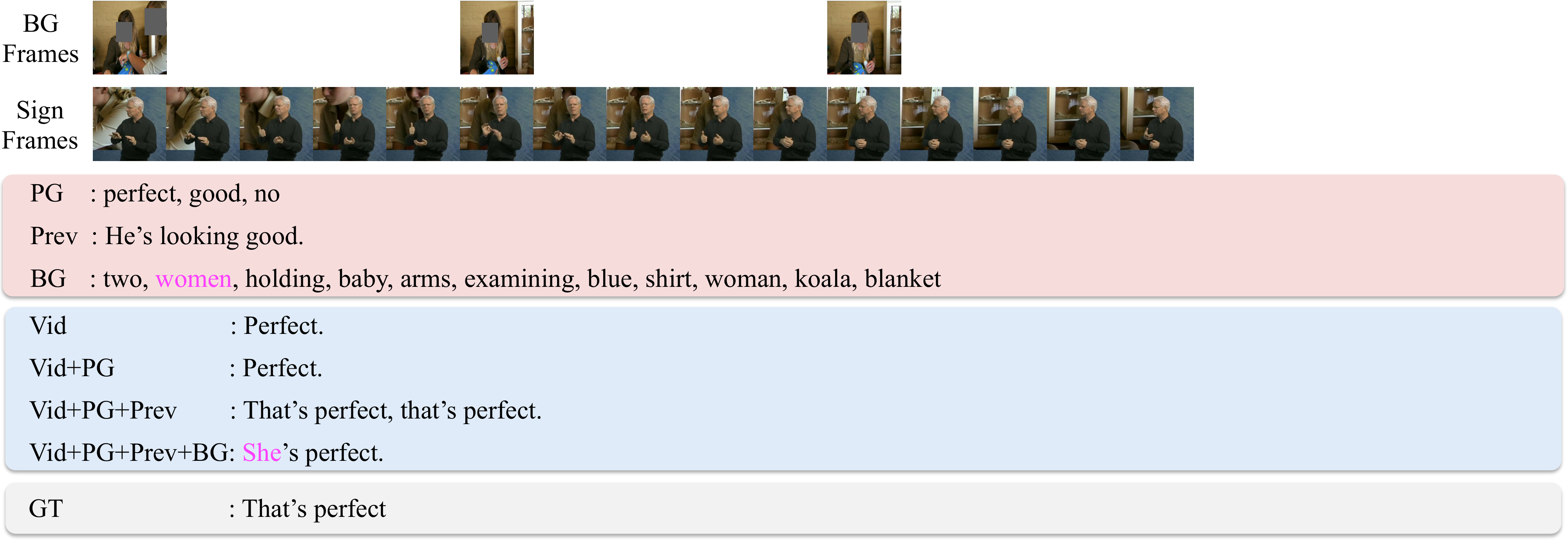}
    \caption{
        \textbf{Qualitative results continued}
    }
    \label{fig:app:qualitative2}
\end{figure*}

We present additional qualitative results 
similar to
\appendixref{Fig.~3}{\cref{fig:qualitative}}
of the main paper, where we display various inputs, and
predictions from our final model compared to baselines.
In the first sample of \cref{fig:app:qualitative1}, the previous sentence indirectly provides information related to location and area, allowing the model to successfully translate the word `river'.
The second sample in \cref{fig:app:qualitative1} demonstrates how the background description conveys information about the presence of multiple people on the screen. 
The third sample in \cref{fig:app:qualitative1} demonstrates the ability of the background description %
to recognise characters on the screen.
The first sample in \cref{fig:app:qualitative2} shows the model accurately capturing the object of the sentence from the background description. The second sample in \cref{fig:app:qualitative2} demonstrates that the models (Vid+PG, Vid+PG+Prev) make incorrect translations by referring to the `cliff' word of the pseudo-glosses. However, when all cues are provided, the bias introduced by the pseudo-glosses is resolved. The third sample in \cref{fig:app:qualitative2} shows a failure case where the model is biased by background information during the process of pronoun assignment.

\noindent\textbf{Supplementary video.}
We provide a supplementary video on our project page showcasing several examples comparing against the baselines, along with dynamic sign language videos. The video further includes an example of continuous translation, where consecutive sign language sentences are translated.

%% file: figs/supmat/llmeval.tex
\begin{figure*}
	\centering
	\begin{tcolorbox}[fontupper=\footnotesize,boxrule=0.5pt,left=1mm,right=1mm,top=1mm,bottom=1mm]
	\{
	
		\quad \textcolor{blue}{``role'': ``system'',}
		
		\quad ``content'': ``Evaluate how well the candidate sentence aligns with the content and meaning of the reference sentence on a scale of 0 to 5.
		
		\qquad \qquad \quad \hspace{1mm} Prioritize key nouns and verbs, while giving less importance to subject, pronouns, adjectives, and adverbs.
		
		\qquad \qquad \quad \hspace{1mm} Scoring Rules:
		
		\qquad \qquad \quad \hspace{1mm} Score at least 1: If the candidate sentence shares at least one key noun or verb (or their synonyms) with the reference sentence.
		
		\qquad \qquad \quad \hspace{1mm} Score at least 3: If the candidate sentence matches most of the key nouns and verbs (or their synonyms) from the reference sentence.
		
		\qquad \qquad \quad \hspace{1mm} Score at least 5: If the candidate sentence conveys the same overall meaning as the reference sentence, with only minor differences.
		
		\qquad \qquad \quad \hspace{1mm} Note: Do not penalize differences in less important words or variations in sentence structure.
		
		\qquad \qquad \quad \hspace{1mm} Focus solely on the essential meaning conveyed by the key nouns and verbs.
		
		\qquad \qquad \quad \hspace{1mm} The candidate sentences are sign language translations of a signer signing the reference sentence.
		
		\qquad \qquad \quad \hspace{1mm} Try to be liberal in the nouns and verbs you consider."

	\},
	
	\textcolor{Green}{\# Example 1}
	
	\{
	
		\quad \textcolor{blue}{``role'': ``user'',}
		
		\quad ``content'': ``\textcolor{Brown}{Assign a score from 0 to 5 based on the rules provided.}
		
		\qquad \qquad \quad \hspace{1mm} \textcolor{Brown}{Provide your answer in JSON format with keys ``score'' (0-5) and ``reason'' with a brief explanation.}
		
		\qquad \qquad \quad \hspace{1mm} \textcolor{Brown}{DO NOT PROVIDE ANY OTHER OUTPUT TEXT OR EXPLANATION. Only provide the JSON string.}
		
		\qquad \qquad \quad \hspace{1mm} \textcolor{Brown}{Reference Sentence:} It's blind to the genius loci.
		
		\qquad \qquad \quad \hspace{1mm} \textcolor{Brown}{Candidate Sentence:} And that's what it means to be dislocated.''
		
	\},

	\{
	
		\quad \textcolor{blue}{``role'': ``assistant'',}
		
		\quad ``content'': ``\{
			
		\qquad \qquad \qquad ``score'': 0,
			
		\qquad \qquad \qquad ``reason'': ``No shared key nouns or verbs; the reference mentions `blind' and `genius loci', while the candidate mentions `dislocated'; meanings
		
		\qquad \qquad \qquad \qquad \qquad are different.''
	
	\qquad \qquad \qquad \}''
		
	\},
	
	\textcolor{Green}{\# Examples continued...}
	
	\{
	
		\quad \textcolor{blue}{``role'': ``user'',}
		
		\quad ``content'': ``\textcolor{Brown}{Assign a score from 0 to 5 based on the rules provided.}
		
		\qquad \qquad \quad \hspace{1mm} \textcolor{Brown}{Provide your answer in JSON format with keys ``score'' (0-5) and ``reason'' with a brief explanation.}
		
		\qquad \qquad \quad \hspace{1mm}  \textcolor{Brown}{DO NOT PROVIDE ANY OTHER OUTPUT TEXT OR EXPLANATION. Only provide the JSON string.}
		
		\qquad \qquad \quad \hspace{1mm}  \textcolor{Brown}{Reference Sentence:}  \{\texttt{text\_gt\}}
		
		\qquad \qquad \quad \hspace{1mm}  \textcolor{Brown}{Candidate Sentence:}  \{\texttt{text\_pred\}}''
		
	\}
	\end{tcolorbox}
	
	\vspace{-.5cm}
	\caption{\textbf{LLM evaluation prompt:} We provide the input format that we feed to \texttt{GPT-4o-mini} \cite{gpt4} to evaluate the quality of the translated sentence (\texttt{text\_pred}) by asking the LLM to compare it against the ground truth sentence (\texttt{text\_gt}). Specifically, we design a system prompt to define the task, and a series of user-assistant prompt pairs to provide input-output examples for calibration. The last user prompt includes the translated sentence to be evaluated. Instructions are repeated at each user prompt.
	Here, we display only one example (enclosed in between \# comment lines to facilitate the reading).
	In practice, we provide 12 in-context examples, which are listed in \cref{tab:app:incontext}, 
	and the full prompt can be found in the code release.
	}
	\label{fig:app:llmevalprompt}
\end{figure*}